\documentclass[
dvipsnames, table,   
format=sigconf,
anonymous=false,     
review=false,        
authorversion=false, 
screen=true, 
nonacm=true, 
]{acmart}

\usepackage{custom_commands}
\usepackage{custom_aliases}
\usepackage{graphics}
\usepackage{bm}
\usepackage{breqn}
\usepackage{setspace}
\usepackage[thinc]{esdiff}
\usepackage{pdfpages}

\newcommand{\mathsym}[1]{{}}
\newcommand{\unicode}[1]{{}}

\newcommand{\Rea}{\mathbb{R}}

\copyrightyear{2022}
\acmYear{2022}
\setcopyright{acmlicensed}\acmConference[GECCO '22 Companion]{Genetic and Evolutionary Computation Conference Companion}{July 9--13, 2022}{Boston, MA, USA}
\acmBooktitle{Genetic and Evolutionary Computation Conference Companion (GECCO '22 Companion), July 9--13, 2022, Boston, MA, USA}
\acmPrice{15.00}
\acmDOI{10.1145/3520304.3533973}
\acmISBN{978-1-4503-9268-6/22/07}

\begin{document}


\title{R-MBO: A Multi-surrogate Approach for Preference Incorporation in Multi-objective Bayesian Optimisation}


\author{Tinkle Chugh}
\email{t.chugh@exeter.ac.uk}
\orcid{0000-0001-5123-8148}
\affiliation{%
  \department{Department of Computer Science}
  \institution{University of Exeter}
  \city{Exeter}
  \country{United Kingdom}
}


\renewcommand{\shortauthors}{Chugh}
\begin{abstract}
Many real-world multi-objective optimisation problems rely on computationally expensive function evaluations. Multi-objective Bayesian optimisation (BO) can be used to alleviate the computation time to find an approximated set of Pareto optimal solutions. In many real-world problems, a decision-maker has some preferences on the objective functions. One approach to incorporate the preferences in multi-objective BO is to use a scalarising function and build a single surrogate model (mono-surrogate approach) on it. This approach has two major limitations. Firstly, the fitness landscape of the scalarising function and the objective functions may not be similar. Secondly, the approach assumes that the scalarising function distribution is Gaussian, and thus a closed-form expression of an acquisition function e.g., expected improvement can be used. We overcome these limitations by building independent surrogate models (multi-surrogate approach) on each objective function and show that distribution of the scalarising function is not Gaussian. We approximate the distribution using Generalised value distribution. We present an a-priori multi-surrogate approach to incorporate the desirable objective function values (or reference point) as the preferences of a decision-maker in multi-objective BO. The results and comparison with existing mono-surrogate approach on benchmark and real-world optimisation problems show the potential of the proposed approach.

\end{abstract}

%
%



\maketitle


\section{Introduction}
Many real-world optimisation problems involve multiple conflicting objectives to be achieved. These problems are called multi-objective optimisation problems (MOPs). There is no single solution to such problems because of the conflicting nature between the objectives. 
The solutions to such problems are known as Pareto optimal solutions, representing the trade-off between objectives \cite{Miettinen1999}. We define a multi-objective optimisation problem (MOP) as:
\begin{equation*}
  \text{minimise} \enspace \bff =   \left(f_1(\bx),\dotsc,f_m(\bx)\right)\qquad\mbox{subject to} \quad \bx \in S 
\end{equation*}
%
with $m \geq 2$ objective functions $f_i(\bx)\colon S\to\Rea$. 
The (nonempty) feasible space $S$ is a subset of the decision space $\Rea^n$ and consists of decision vectors $\bx=(x_1,\ldots,x_n)^T$. There are three methods to solve such problems based on the preferences of a decision-maker (DM): A priori, A posteriori and Interactive \cite{Miettinen1999}. This work focuses on a prior approach that aims to find a single solution preferable to the DM.

In many MOPs the objective function rely on computationally expensive evaluations. Such problems are usually black-box optimisation problems without any closed-form for the objective functions. Multi-objective Bayesian optimisation (BO) can be used to alleviate the computational cost and to find an approximated set of Pareto optimal solution(s) in the least number of function evaluations. These methods rely on a Bayesian model as the surrogate (or metamodel) of the objective functions and find promising decision vectors by optimising an acquisition function. The Bayesian model is usually a Gaussian process because it provides a meaningful quantification of uncertainty, which is then used in optimising the acquisition function. The acquisition function balances both exploration and exploitation in guiding the search process. 


In multi-objective BO, there are typically two different approaches to build a Bayesian model. In the first one, the models are built for each objective function and an acquisition function utilising these models is then used to find promising decision vectors. This approach is called multi-surrogate approach. The multi-objective BO with expected hypervolume improvement (EHVI) \cite{Emmerich2004,Emmerich2005a,Emmerich2006,Emmerich2011,Yang2019} is a multi-surrogate approach. In the second one, a single Bayesian model is built after aggregating the objective functions. This approach is called mono-surrogate approach. The well-known ParEGO \cite{Knowles2006} algorithm comes under the second category. The second approach reduces the number of objectives from $m$ to one. Moreover, a single objective acquisition function can be used in the mono-surrogate approach. The computational complexity of the first approach is at most $O(mN^3)$ and of the second approach is at most $O(N^3)$, where $N$ is the size of the data set. 

The ParEGO converts the multiple objectives into a single objective by utilising the augmented weighted Tchebycheff (TCH) as the scalarising function. A Gaussian process model is then built on the scalarising function, which is then used in optimising the expected improvement (EI) to find the next promising decision vector. In a recent work \cite{Hakanen2017}, ParEGO was extend to handle the reference point (or desirable objective function value) as the DM's preferences. The algorithm used Achievement scalarising function (ASF) instead of TCH and built a Gaussian process model on it. The ASF uses the reference point instead of ideal objective vector and has been widely used in the multiple criteria decision making literature \cite{Miettinen1999}. Although, the mono-surrogate approach uses only one model, it has two major limitations. The first one is that the fitness landscape of the scalarising function and the objective functions may not be similar. In other words, a promising decision vector by using the surrogate on the scalarising function may not be promising for the underlying objective functions. The second limitation is that the approach assumes that the resulting scalarising function is Gaussian and thus a closed-form expression of the EI can be used.

In this work, we overcome the limitations of mono-surrogate approach by building independent models on the objective functions. We embed the DM's preferences into the ASF and show that the distribution of ASF after building independent models is not Gaussian. Therefore, we approximate the distribution of the scalarising function using the Generalised extreme value (GEV) distribution \cite{GEV,Haan2006}. Particularly, we use the type I family of distributions in the generalised extreme value theory to approximate the distribution. The GEV distribution is then used in optimising the expected improvement to find the next promising decision vector. The proposed approach is called R-MBO and handles the preferences as a-priori.


We compare the proposed multi-surrogate EI with mono-surrogate EI \cite{Hakanen2017} on standard benchmark and real-world optimisation problems. We generate arbitrary reference points and assume that they are the DM's desirable objective function values. The results on benchmark and real-world multi-objective optimisation problems clearly show the potential of the proposed work. 

The rest of the paper is structured as follows. In Section 2, we provide the background on BO and overview of the mono-surrogate approach i.e.\ ParEGO with DM's preferences \cite{Hakanen2017}. In Section 3, we explain the proposed approach by comparing it with the mono-surrogate approach. In Section 4, we conduct numerical experiments and show the results on benchmark and real-world multi-objective optimisation problems. Finally, we conclude and mention the future research direction in Section 5. 

\section{Bayesian Optimisation}
In multi-objective BO, the input is the data set $\{ (\bx_i, \bff(\bx_i))\}_{i=1}^N$ of size $N$. This data set can be obtained with a design of experiment technique e.g.\ Latin Hypercube sampling \cite{Mckay2000}. 
The Gaussian process models are the most commonly used Bayesian models in BO. They are non-parametric and provide uncertainties in predictions, which makes them different from other non-Bayesian and parametric models. The uncertainty is then used in the acquisition function in finding the promising decision vector. A Gaussian process ($\mGP$) can be defined with a multivariate normal distribution \cite{Rasmussen2006}:
\begin{align*}
  \mathbf{f} \sim \mathcal{N}(\bm{\mu}, K),
\end{align*}
where $\bm{\mu}$ is the mean vector and $K$ is the covariance matrix. Without loss of generality, we use a prior of zero mean. A covariance function (or kernel) is used to get the covariance matrix. In this work, we used a Gaussian (or RBF or squared exponential) kernel with automatic relevant determination \cite{Chugh_2019_LOD,Palar_AIAA}:
\begin{align*}
    \kappa(\mathbf{x}, \mathbf{x'}, \mathbf{\Theta}) = \sigma_f^2
    \exp\left(-\frac{1}{2}\sum\limits_{j=1}^n\frac{|x_{j} - x_{j}'|^2}{l_j^2}\right) + \sigma_n^2 \delta_{\mathbf{x}\mathbf{x'}},
\end{align*} 
where $\Theta = (\sigma_f, l_1,\ldots,l_n,\sigma_n)$ is the set of hyperparameters and $\delta_{\mathbf{x}\mathbf{x'}}$ is the Kronecker delta function. The notation $|x_{j} - x_{j}'|$ represents the Euclidean distance between $x_j$ and $x_j'$. The hyperparameters can be estimated by maximising the marginal likelihood function \cite{Rasmussen2006}:
\begin{align*}
    p(\mathbf{f}|X,\mathbf{\Theta}) = \frac{1}{|2\pi K |^{\frac{1}{2}}} \exp \Big(
  -\tfrac{1}{2} \mathbf{f}^T K^{-1} \mathbf{f}
  \Big).
\end{align*}
The posterior predictive distribution at new point $\bx^*$after training the model is also Gaussian:

\begin{align*}
\begin{split}
 p\left(f^* | \mathbf{x^*},X,\mathbf{f},\mathbf{\Theta} \right) =
 \mathcal{N}\Big( \bm{\kappa}(\mathbf{x^*}, X) K ^{-1} \mathbf{f},
 \kappa(\mathbf{x^*}, \mathbf{x^*}) - \bm{\kappa}(\mathbf{x^*}, X) ^T K ^{-1} \bm{\kappa}(X, \mathbf{x^*}) \Big).
 \end{split}
\end{align*}



%

The acquisition function determines the next decision vector to be evaluated. The expected improvement (EI) is one of the well-known and widely used acquisition functions. It measures the amount of improvement over the current best objective value and balances both exploitation and exploration. For a minimisation problem, the improvement over the best evaluated function value $f'(\bx)$ is:
\begin{align*}
    I(\bx) = \max (0, f'(\bx) - f).
\end{align*}
The expected improvement can then be estimated as:
\begin{align*}
    \alpha_{EI}(\bx) = \int_{-\infty}^{f'(\bx)}I(x) df
\end{align*}
As the posterior is Gaussian, the expected improvement has a closed-form expression:
\begin{align*}
    \alpha_{EI}(\bx) = (f'(\bx) - \mu(\bx))\Phi\Big(\frac{f'(\bx) - \mu(\bx)}{\sigma(\bx)}\Big) + \sigma (\bx) \phi\Big(\frac{f'(\bx) - \mu(\bx)}{\sigma(\bx)}\Big),
\end{align*}
where $\mu(x)$ and $\sigma(x)$ is the posterior mean and standard deviation, respectively and $\Phi(\cdot)$ and $\phi(\cdot)$ are cumulative and probability distribution function of standard normal distribution, respectively.

In mono-surrogate multi-objective BO, this formulation of EI can be used as the objective functions are aggregated into a single function. In the multi-surrogate approach, EHVI, which is an extension of EI for multiple objectives can be used. In this work, we focus on EI after using multiple surrogates. The EI is usually a multimodal function and therefore, a suitable optimiser e.g.\ an evolutionary algorithm is often used to optimise it to find the next decision vector, which is then evaluated with expensive objective function and added to the data set. This process continues until a termination criterion is met. Algorithm \ref{alg:BO} outlines these steps.



\begin{algorithm}
\caption{Bayesian optimisation}
\label{alg:BO}
\begin{algorithmic}[]
    \State \textbf{Input}: 
        Data Set $D = \{ (\bx_i,\bff_i) \}_{i=1}^N$
    \State \textbf{Output}: 
        Evaluated solutions
\end{algorithmic}
\begin{algorithmic}[1]
    \While{Termination criterion is not met}
    \State Train the $\mGP$ models on the data set
    \State Optimise the acquisition function i.e.\ $\bx^* \leftarrow \argmax_{\bx} EI(x)$ 
    \State Evaluate $\bx^*$ and add to the data set
    \EndWhile
\end{algorithmic}
\end{algorithm}

\subsection{Preference incorporation in Mono-surrogate approach}
The mono-surrogate approach is one of the well-known approaches in multi-objective BO. The ParEGO algorithm is a mono-surrogate approach and uses the weighted Tchebycheff (TCH) for scalarising the objective functions. The TCH function is defined as:
\begin{align*}
    g = \max_{i} \big( w_i (f_i - z_i)\big),
\end{align*}
where $\bww$ is the weight vector and $\mathbf{z}$ is the ideal objective vector (or minimum of objective function values). In ParEGO, an augmented Tchebycheff formulation was used to obtain properly Pareto optimal solutions \cite{Miettinen1999}. If the objective function values at the current iteration are normalised between 0 and 1, then $\mathbf{z}$ is a vector of zeros. Given a data set with decision variable and objective function values, the TCH converts the multiple objective function values into a single value. A Gaussian process model is then built on the resulting data set, which is then used in optimising the expected improvement. Recently, three other scalarising functions called hypervolume improvement, dominance ranking and sign distance were proposed in \cite{Alma2017} and used in the framework of the mono-surrogate approach.


In \cite{Hakanen2017}, the ParEGO algorithm was extended to incorporate the decision-maker's preferences. The algorithm was used to handle the preference iteratively during the solution process. However, the algorithm can be used as a-priori. The only change from the original ParEGO was the use of Achievement scalarising function (ASF) \cite{Wierzbicki1980} instead of weighted Tchebycheff. The ASF is defined as:
\begin{align*}
    g = \max_{i} \big( w_i (f_i - z_i^*)\big),
\end{align*}
where $\mathbf{z}^*$ is the desirable objective function vector. The ASF has been widely used in Multiple Criteria Decision Making \cite{Miettinen1999, Miettinen2002, Miettinen2008, Miettinen2016} because of its ability to handle DM's preferences. As suggested in \cite{Miettinen1999}, the authors in \cite{Hakanen2017} used $w_i = \frac{1}{z_i^{nadir} - z_i^{ideal}}$ to normalise the objective function values. The $\mathbf{z}^{nadir}$ and $\mathbf{z}^{ideal}$ are nadir and ideal objective vector \cite{Deb2010}, respectively. If these vectors are not available, the maximum and minimum of the objective function values at the current iteration can be used. After building a Gaussian process model on the ASF, the algorithm maximised the EI to find the next decision vector. In this way, the algorithm aimed to find a single solution preferable to the DM.



\section{R-MBO: Multi-surrogate Approach }
\begin{figure*}
    \centering
    \includegraphics[width=0.33\textwidth]{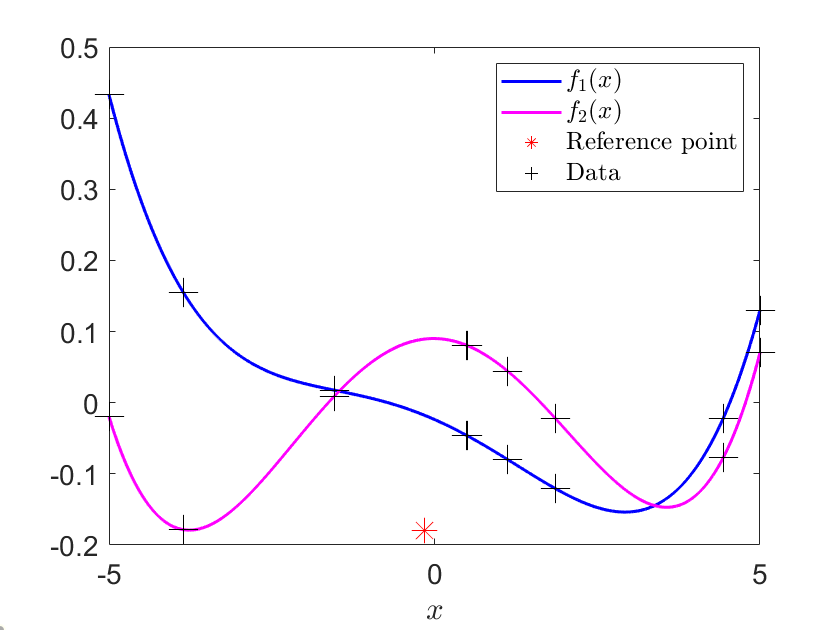}
    \includegraphics[width=0.33\textwidth]{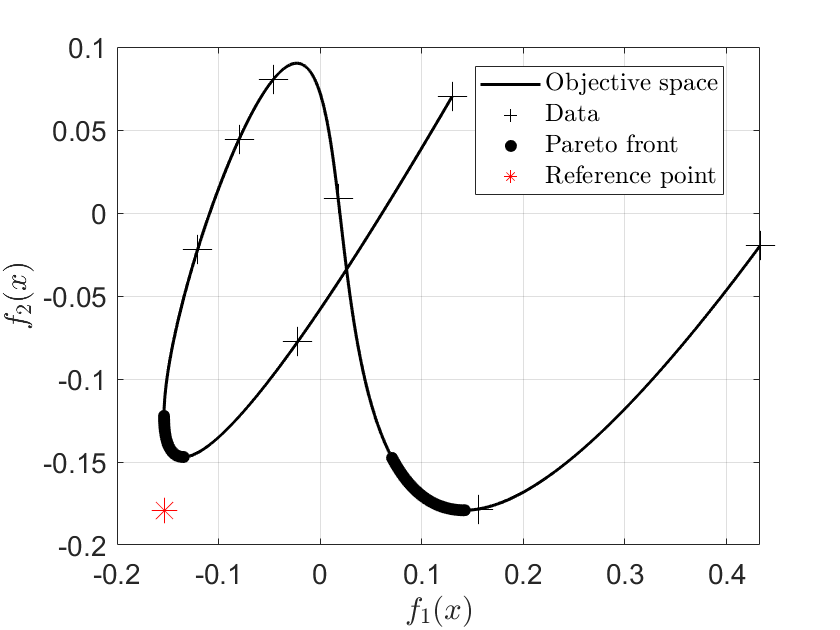}
    \includegraphics[width=0.33\textwidth]{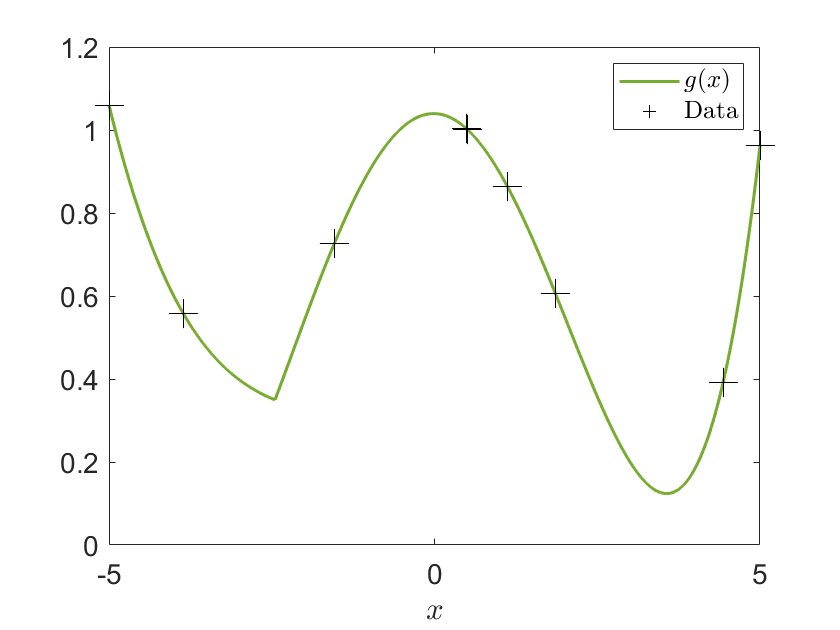}
    \caption{A bi-objective optimisation problem. Both objectives are to be minimised. The data is shown in '+' in the left figure. The Pareto front and the objective space are shown in the middle figure. The resulting achievement scalarising function with the data set is shown in the right figure.}
    \label{fig:demo_1}
\end{figure*}

\begin{figure*}
    \centering
    \includegraphics[width=0.45\textwidth]{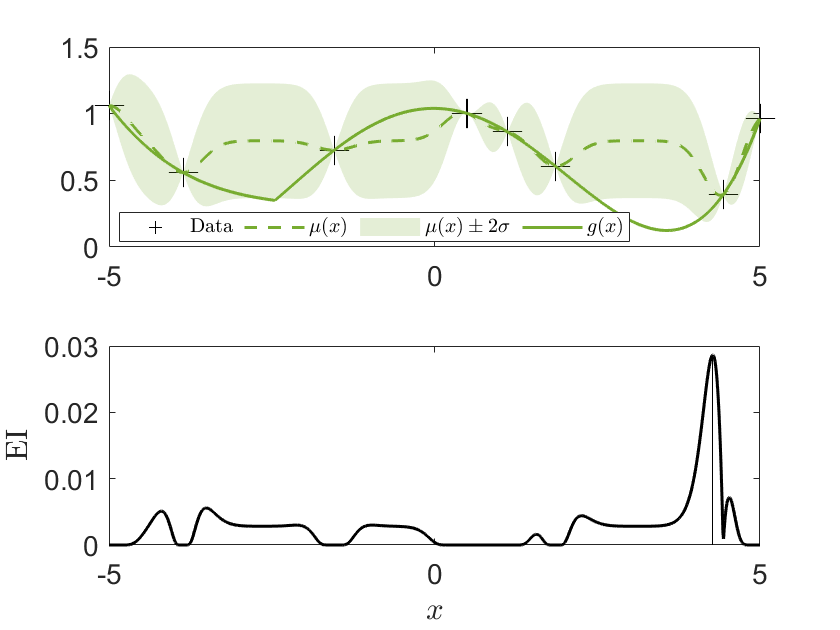}
    \includegraphics[width=0.45\textwidth]{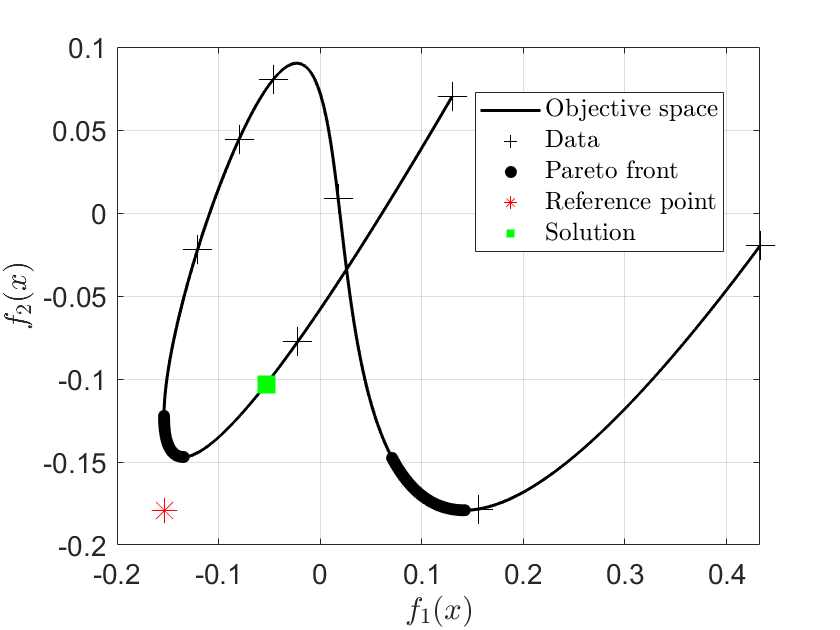}
    \caption{A Gaussian process model and the landscape of EI on the achievement scalarising function (left figure). The evaluated solution (shown as the square marker) in the objective space after maximising the EI (right figure).}
    \label{fig:demo_2}
\end{figure*}

In the multi-surrogate approach, we build independent Gaussian process models on the objective functions. These independent models are then used to build a probabilistic model for the ASF, which is then used in optimising the EI. We start by providing a simple example to compare mono and multi surrogate approaches with ASF as the scalarising function. Consider two objective functions $f_1$ and $f_2$ shown in Figure \ref{fig:demo_1}. We assume that these functions are black box and their analytical forms are not available. However, we have some data set (with decision variable and objective functions values) shown as $+$ in the figure. We assume that the ideal and nadir objective vectors are available. For a reference point (or desirable objective function vector) i.e.\ $\mathbf{z^*} = \mathbf{z^{ideal}}$, the true or underlying ASF (denoted as $g(x)$) is also shown in the figure. As mentioned, there are two ways to build surrogate models: mono-surrogate and multi-surrogate. In the mono-surrogate approach, we get the ASF values after aggregating two objective values (right plot in Figure \ref{fig:demo_1}). We then build a Gaussian process model on it as shown in Figure \ref{fig:demo_2}. We then maximise the EI to get the next decision variable value i.e.\ $x^* = \argmax \text{EI}$. The location of $x^*$ can be seen in Figure \ref{fig:demo_2}. We then evaluate the $x^*$ with the objective functions. The resulting objective function values are shown as square in the right plot in Figure \ref{fig:demo_2}. In this way, the mono-surrogate approach tries to solve the single-objective optimisation problem and aims to find a solution preferable to the DM.


\begin{figure*}
    \centering
    \includegraphics[width=0.45\textwidth]{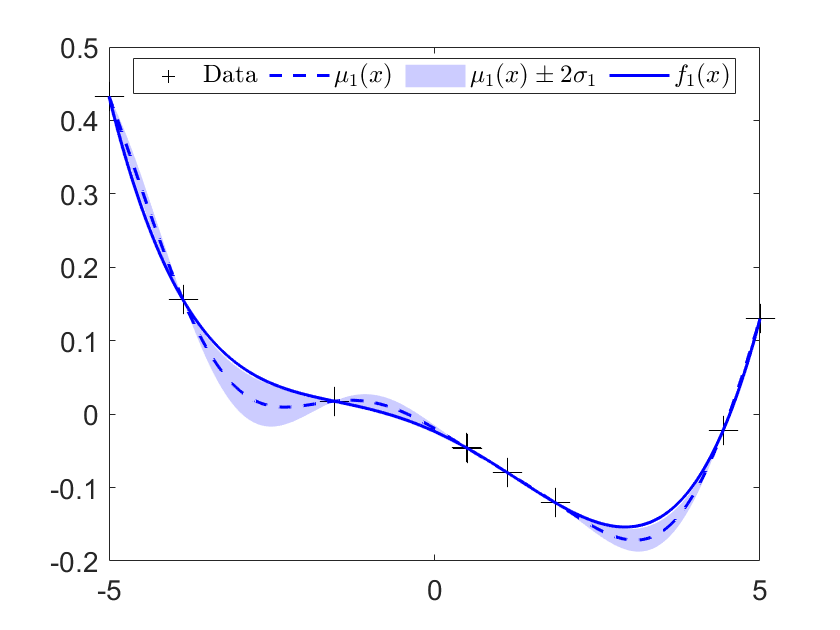}
    \includegraphics[width=0.45\textwidth]{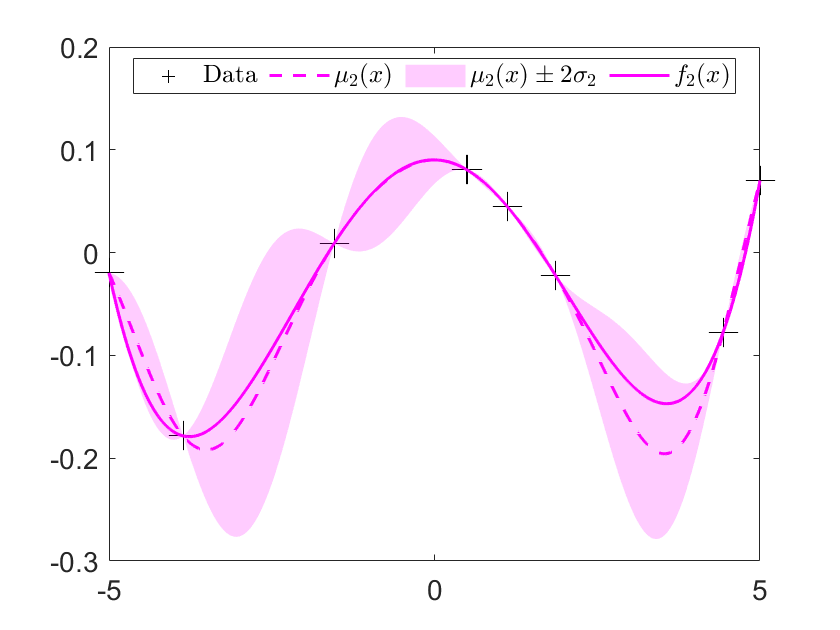}
    \caption{A Gaussian process model for each objective function.}
    \label{fig:demo_3}
\end{figure*}

\begin{figure*}
    \centering
    \includegraphics[width=0.4\textwidth]{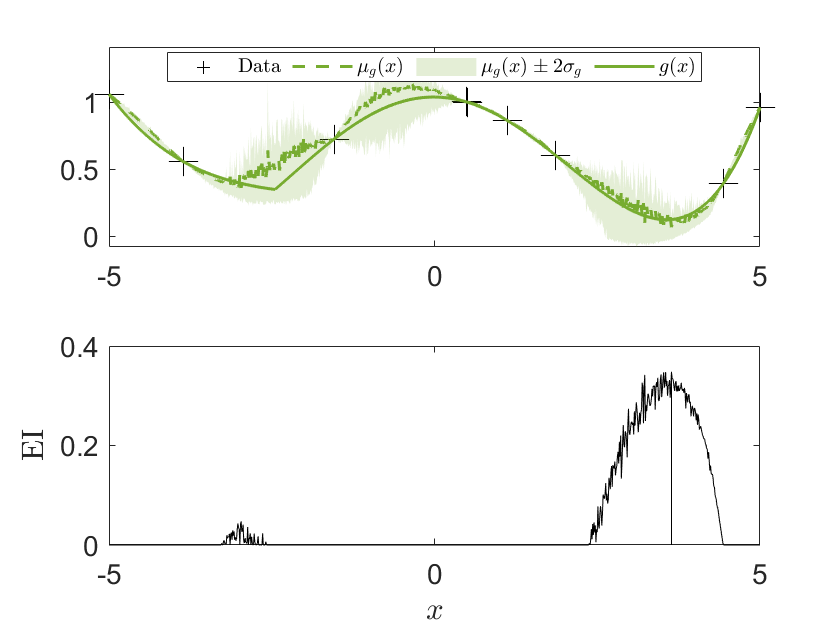}
    \includegraphics[width=0.4\textwidth]{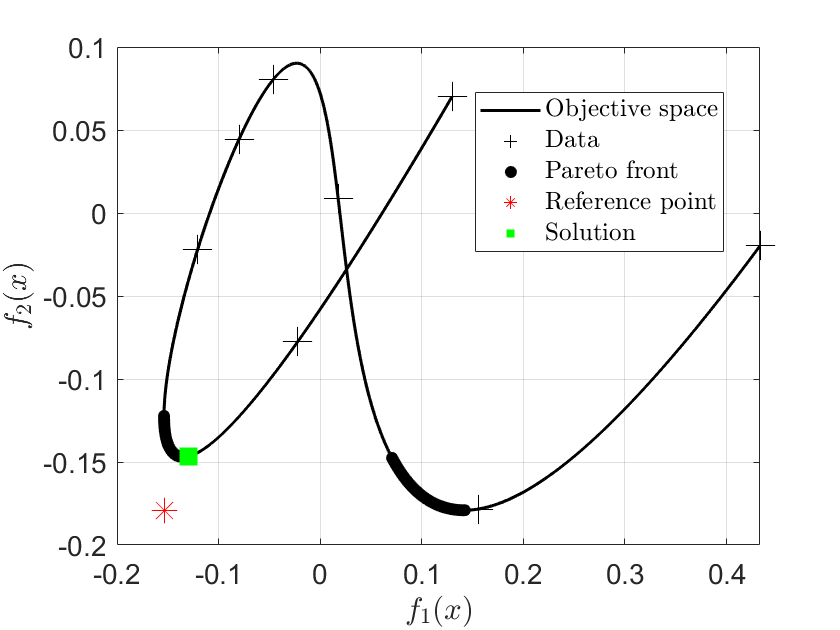}
    \caption{A Gaussian process model and the landscape of EI on the achievement scalarising function after building independent models for each objective function (left figure). The evaluated solution (shown as the square marker) in the objective space after maximising the EI (right figure).}
    \label{fig:demo_4}
\end{figure*}

In multi-surrogate approach, we can build independent models for each objective function. The predictions with uncertainty estimates after building independent models are shown in Figure \ref{fig:demo_3}. The question is how to use these models to find the distribution of the ASF as the scalarising function. We start answering this question by showing that the resulting scalarising function after building independent models on objective functions is not Gaussian. In the ASF function, $g = \max_i\big( w_i(f_i - z_i^*) \big)$, $f_i$ is Gaussian i.e.\ $f_i \sim \mathcal{N}(\mu_i, \sigma_i^2)$, where $\mu_i$ and $\sigma_i$ are the posterior predictive means and standard deviations, respectively. The formulation of $g$ can be written as \cite{Lemons2002,Mood1974}:
\begin{align}
     g \sim \max_i \mathcal{N}\Big( w_i(\mu_i - z_i^*), w_i^2 \sigma_i^2\Big).
     \label{eq:g_pdf1}
\end{align}
After some rearrangements, the distribution shown above can be written in the following closed form expression \cite{Azzalini1985,Nadarajah2008,Atanu_TEVC}: 

\begin{dmath}
    p(g) =\sum_{i=1}^m \frac{1}{w_i\sigma_i} \times \frac{\phi\Big( \frac{g - w_i(\mu_i - z_i^*)}{w_i\sigma_i} \Big)}{\Phi\Big( \frac{g - w_i(\mu_i - z_i^*)}{w_i\sigma_i}\Big)} \times 
     \prod_{i=1}^m \Phi\Big( \frac{g - w_i(\mu_i - z_i^*)}{w_i\sigma_i}\Big),
    \label{eq:g_pdf2}
\end{dmath}
where $\phi(\cdot)$ and $\Phi(\cdot)$ represent the probability density and cumulative density functions of the standard normal distribution, respectively. The distribution in Equation (\ref{eq:g_pdf2}) has a closed form expression but it is not Gaussian distributed. Therefore, a closed form expression of acquisition functions, which rely on the Gaussian assumption of the function cannot be used. A possible solution to this problem is to approximate the distribution with Generalised value theory \cite{GEV,Haan2006}. The distribution can be described with Type 1 distribution in Generalised value theory . Specifically, we use a Gumbel distribution \cite{GumbelBO} for approximating the scalarising function:
\begin{align*}
    g \sim  \text{Gumbel} \ (\alpha,\beta) 
\end{align*}
where $\alpha$ and $\beta$ are location and scale parameters, respectively. The probability density function is:
\begin{align*}
    p(g_i|\alpha,\beta) =& \frac{1}{\beta}e^{-(t_i + e^{-t_i})} \enspace \text{for} \enspace i =1,\ldots,N 
\end{align*}
where $t_i =\frac{g_i - \alpha}{\beta}$. We can estimate the parameters by maximising the following log-likelihood function
\begin{align*}
    LL(\alpha,\beta) = N \log \frac{1}{\beta} - \sum_{i=1}^N t_i - \sum_{i=1}^N e^{-t_i},
\end{align*}
where $N$ is the number of samples drawn from Equation (\ref{eq:g_pdf1}) or (\ref{eq:g_pdf2}).
The estimated parameters are:
\begin{align*}
    \alpha,\beta =&\argmax_{\alpha,\beta}\prod_{i=1}^N p(g_i|\alpha,\beta) \enspace \text{Or} \\ 
    \alpha,\beta =& \argmax_{\alpha,\beta} \Big( N \log \frac{1}{\beta} - \sum_{i=1}^N t_i - \sum_{i=1}^N e^{-t_i} \Big)
\end{align*}
After taking the partial derivatives, the parameters can be estimated by solving the following two equations:
\begin{align*}
    \beta =& \bar{g} - \frac{\sum_{i=1}^N g_i e^{-\frac{g_i}{\beta}}}{\sum_{i=1}^N e^{-\frac{g_i}{\beta}}} \\
    \alpha =& -\beta \log \Big( \frac{1}{N} \sum_{i=1}^N  e^{\frac{-g_i}{\beta}}\Big)
\end{align*}
Once the parameters are known, we can use the approximated distribution in estimating the EI: 
\begin{align*}
    \alpha_{EI} = \int_{-\infty}^{g'(\bx)} \max (0, g'(x) - g) \ dg
\end{align*}
One major drawback is that the EI does not have a closed-form expression. Therefore, we use Monte Carlo for estimating the EI. For the two-objective example, the resulting ASF predictions and uncertainty estimates after building independent Gaussian process models on objective functions are shown in Figure \ref{fig:demo_4}. We also show the landscape of the expected improvement in the figure and can see that the optimal location is different from the mono-surrogate approach. The resulting decision variable is then evaluated with underlying objective functions and as can be seen, the objective function values lie on the Pareto front. This demonstration on an easy one-dimensional two objective optimisation problem shows that the multi-surrogate approach is better and finding an appropriate distribution of the scalarising function is important. 



\section{Numerical Experiments}
\begin{figure}
    \centering
    \includegraphics[height = 5cm, width = 8.5cm]{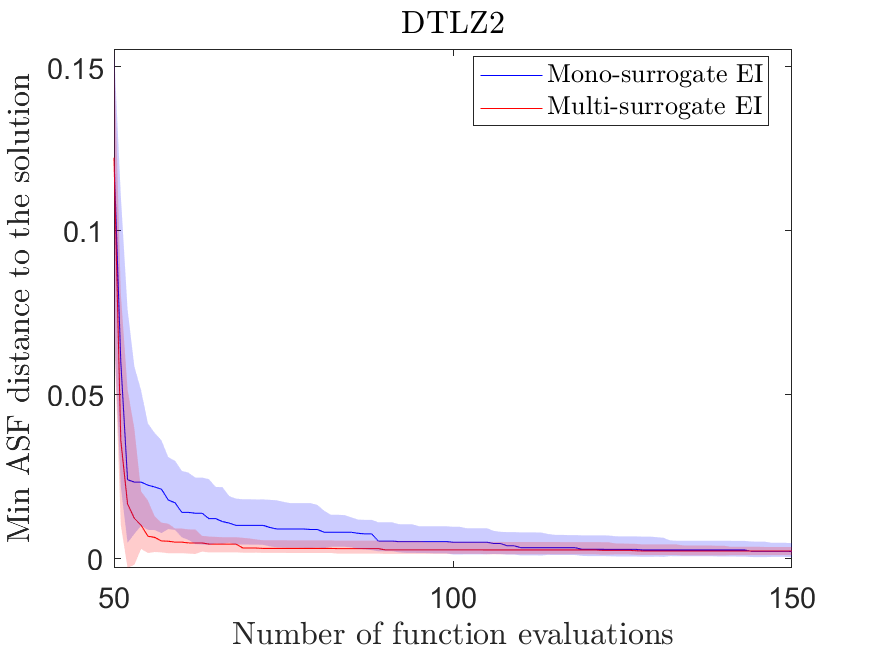}
    \includegraphics[height = 5cm, width = 8.5cm]{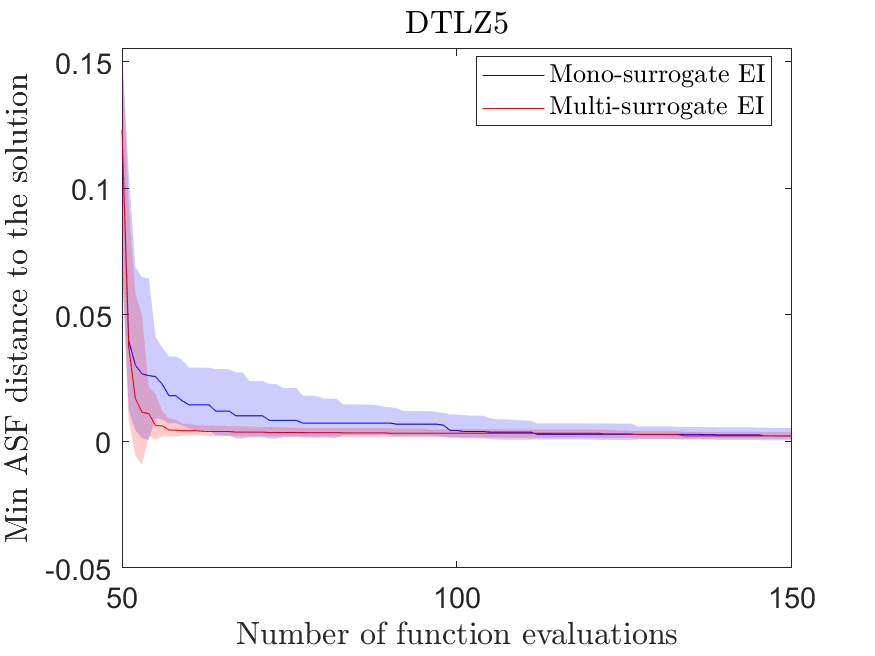}
    \includegraphics[height = 5cm, width = 8.5cm]{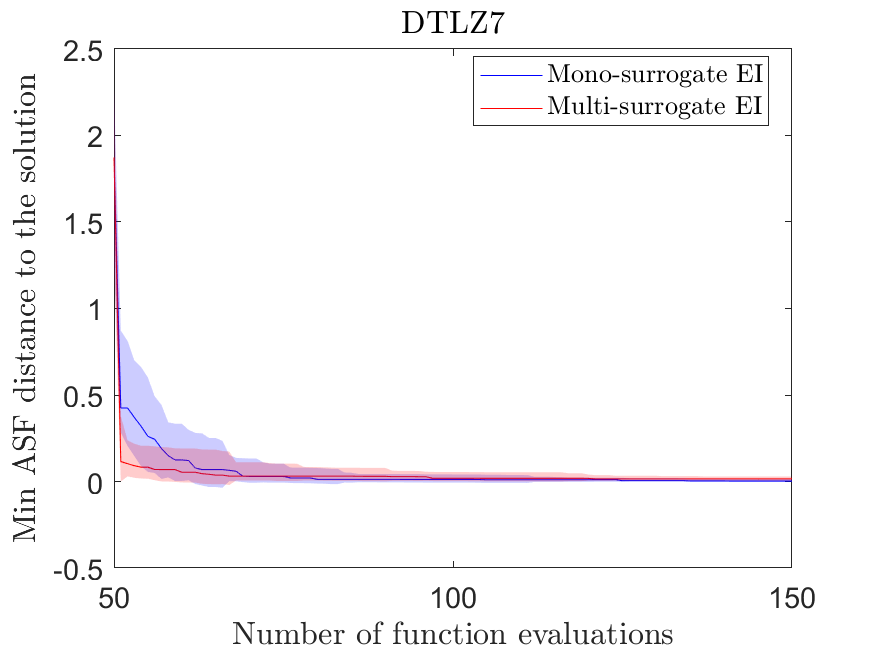}
    \caption{Minimum ASF distance to `Ref Solution' with the number of function evaluations.}
    \label{fig:dtlz2_dist}
\end{figure}

We investigate the performance of the proposed multi-surrogate R-MBO approach on standard DTLZ \cite{Deb2005b} benchmarks and a real-world Free-radical Polymerisation problem \cite{Chugh2014}. We compare it to the mono-surrogate approach \cite{Hakanen2017}. We implemented all different approaches and used the same settings wherever possible for a fair comparison. The numerical settings used are as follows:
\begin{itemize}
    \item Problems: DTLZ (2, 5 and 7), Free-radical Polymerisation
    \item Number of objectives: 2 and 3
    \item Number of decision variables: 4 --5
    \item Number of instances in DTLZ = 25 (number of arbitrary reference point or desirable objective function vectors, generated on a grid with prefixed lower and upper bounds)
    \item Number of instances in Free-radical polymerisation = 1
    \item Number of runs per instance = 1
    \item Size of the initial data set: 10 $\times$ number of decision variables
    \item Maximum number of function evaluations: 30 $\times$ number of decision variables 
    \item Kernel: Squared exponential (or Radial basis function, Gaussian) with automatic relevant determination
    \item Optimiser to maximise acquisition functions: Genetic Algorithm
    \item Optimiser to maximise marginal likelihood in Gaussian process: BFGS with 10 restarts 
    \item Performance indicator: Distance to the solution
    \item Visulisation: Scatter plots
\end{itemize}

To measure the performance of approaches, we sample the Pareto front and find the nearest solution to the reference point by computing: $\bx^* = \argmin_S \max_{i} \big( w_i (f_i(\bx) - z_i^*(\bx))\big)$. As the Pareto fronts of benchmark problems are available, the solution can easily be computed. We denote the nearest solution by `Ref Solution'. For Free radical polymerisation, the Pareto front is not available and therefore, we ran NSGA-II \cite{Deb2000,Deb2002} and used non-dominated solutions as the representation of the Pareto front.  We then compute the ASF distance between solutions of both approaches and `Ref Solution'. As the aim in both approaches is to find a single solution preferable to the DM, the distance to `Ref solution' provides a meaningful indicator as the performance of approaches. Figure \ref{fig:dtlz2_dist} shows the minimum distance between solutions obtained of both approaches and `Ref Solution'. The solid line is the median and shaded region is 68\% confidence interval of 25 independent instances. As can be seen, both approaches converged and found solutions close to `Ref Solution'. However, the multi-surrogate approach converged faster than the mono-surrogate approach.



\begin{figure*}[t]
    \centering
    \includegraphics[width = 0.33\textwidth]{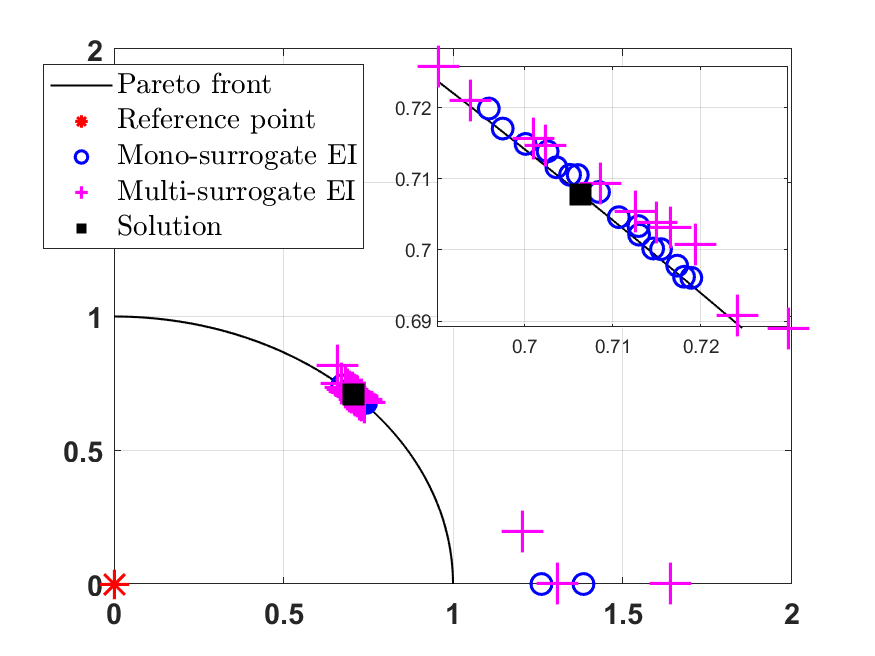}
    \includegraphics[width = 0.33\textwidth]{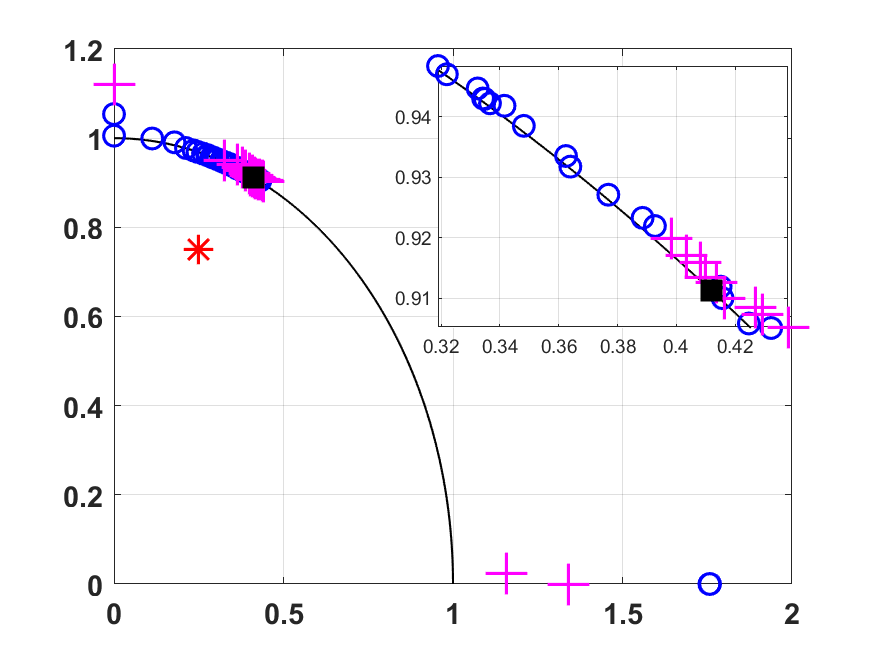}
    \includegraphics[width = 0.33\textwidth]{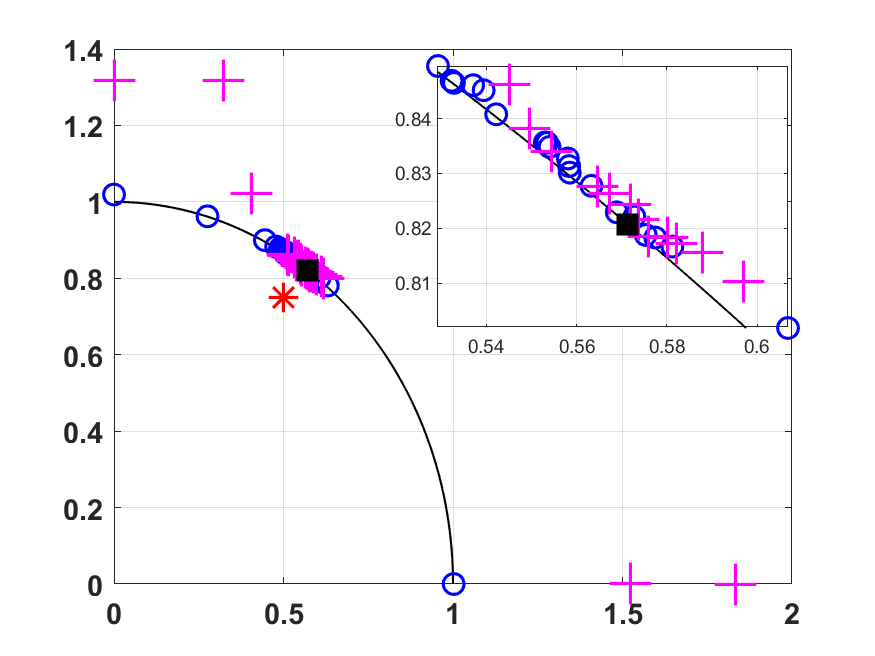}
    \includegraphics[width = 0.33\textwidth]{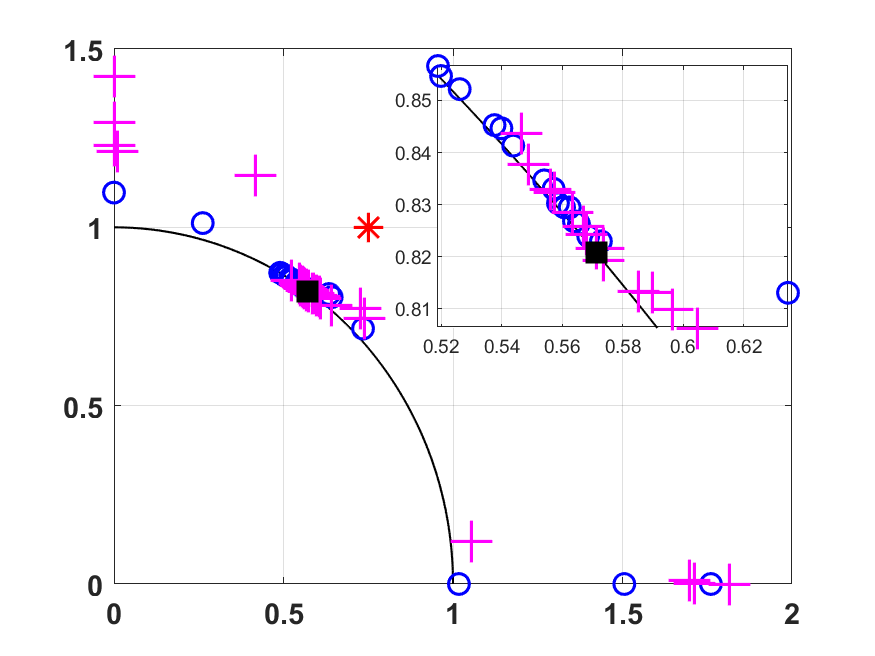}
    \includegraphics[width = 0.33\textwidth]{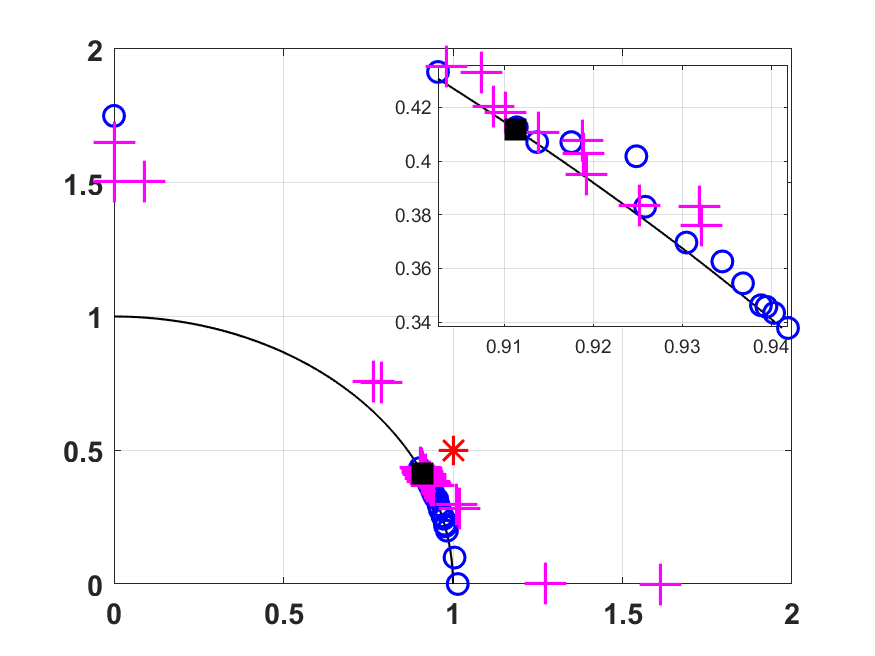}
    \includegraphics[width = 0.33\textwidth]{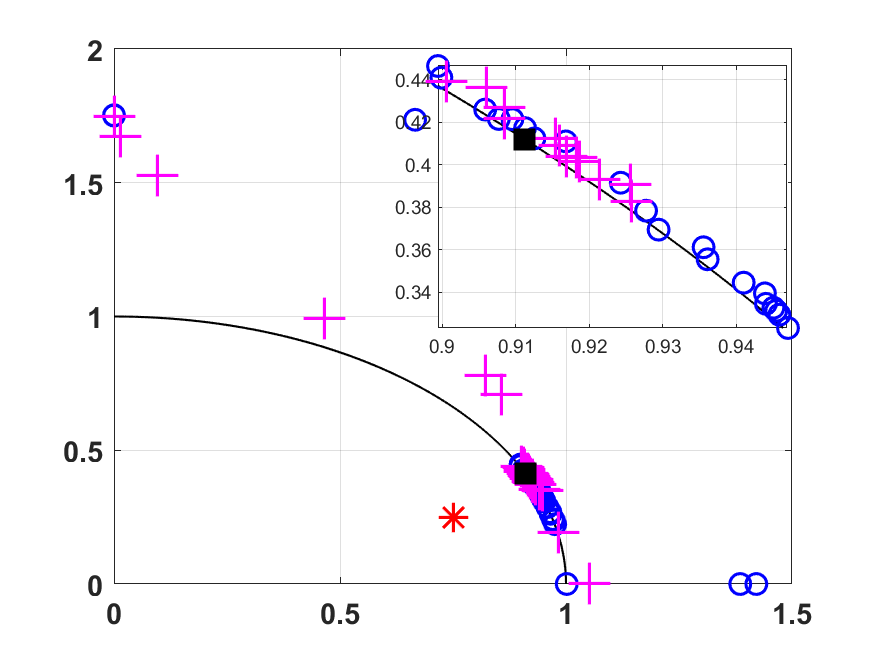}
    \caption{Performance on DTLZ2 problem}
    \label{fig:dtlz2}
\end{figure*}

\begin{figure*}[t]
    \centering
    \includegraphics[width = 0.33\textwidth]{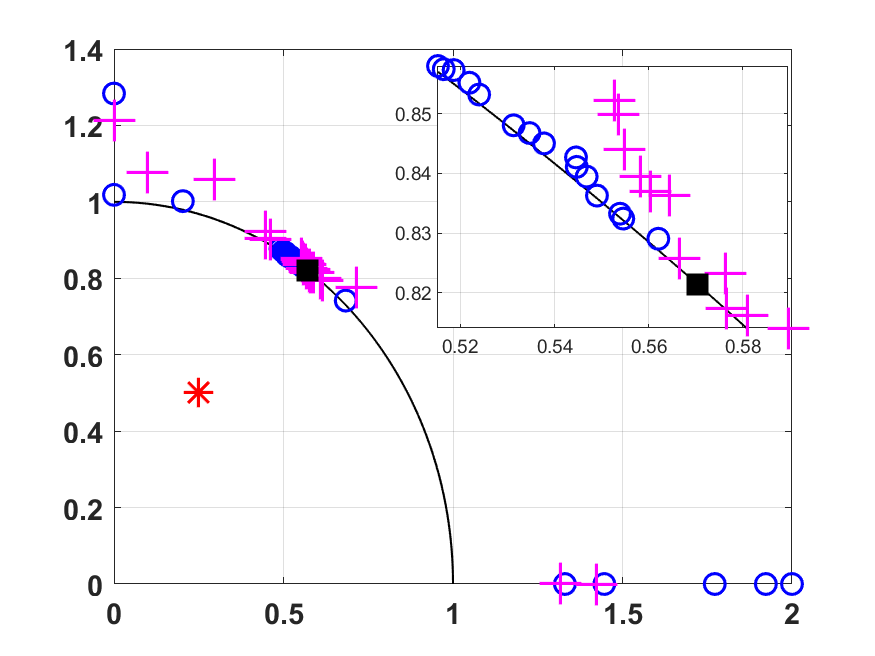}
    \includegraphics[width = 0.33\textwidth]{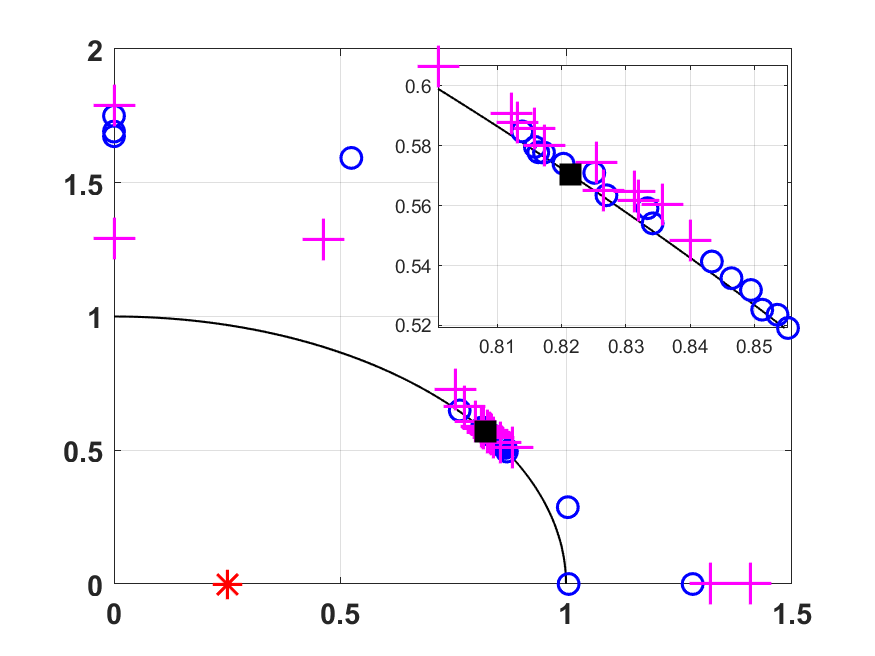}
    \includegraphics[width = 0.33\textwidth]{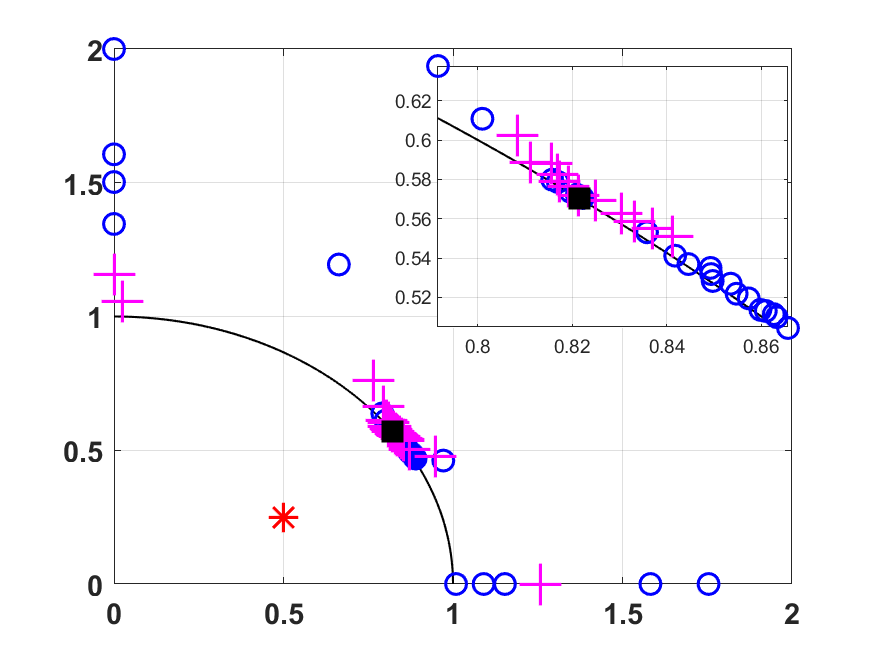}
    \includegraphics[width = 0.33\textwidth]{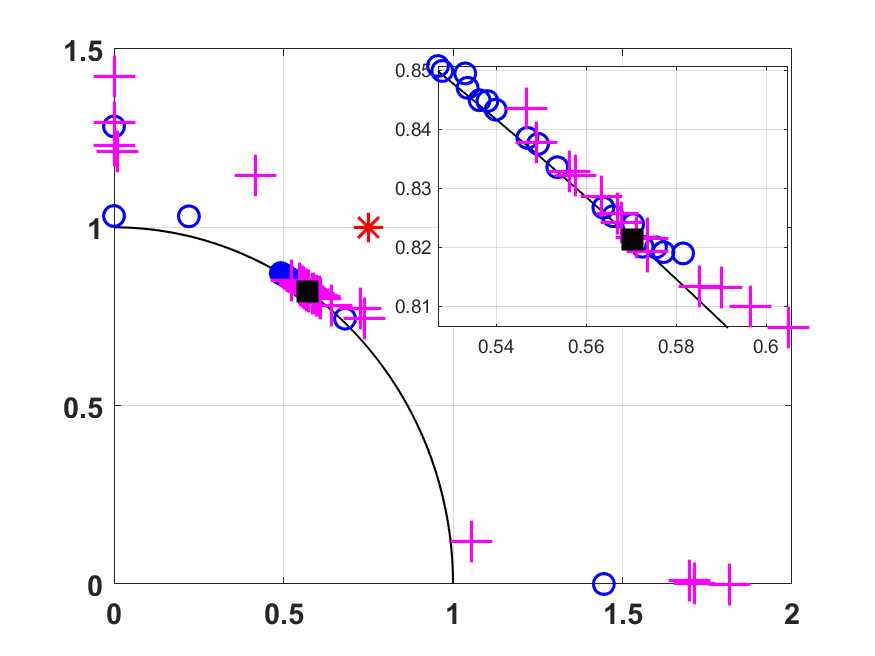}
    \includegraphics[width = 0.33\textwidth]{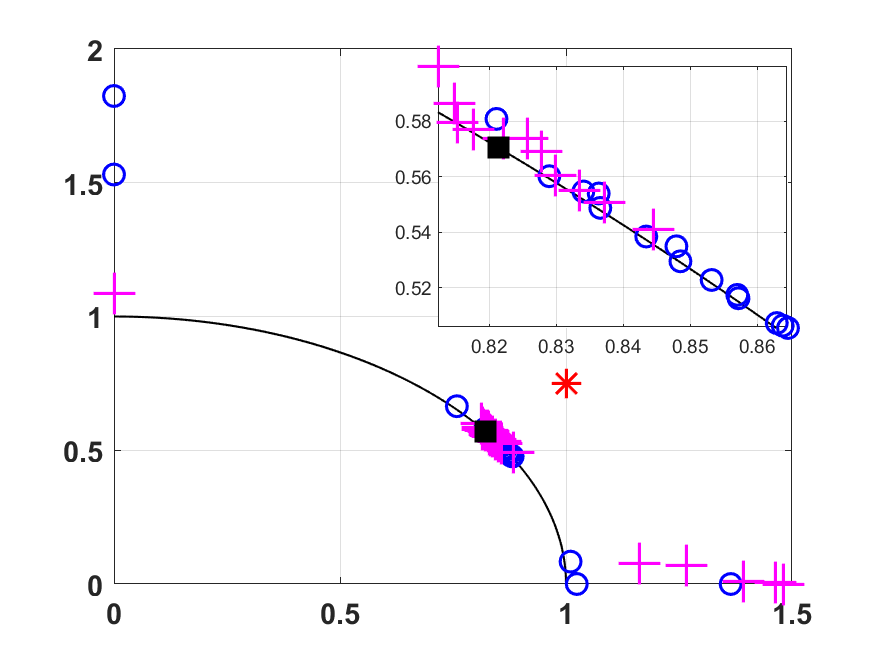}
    \includegraphics[width = 0.33\textwidth]{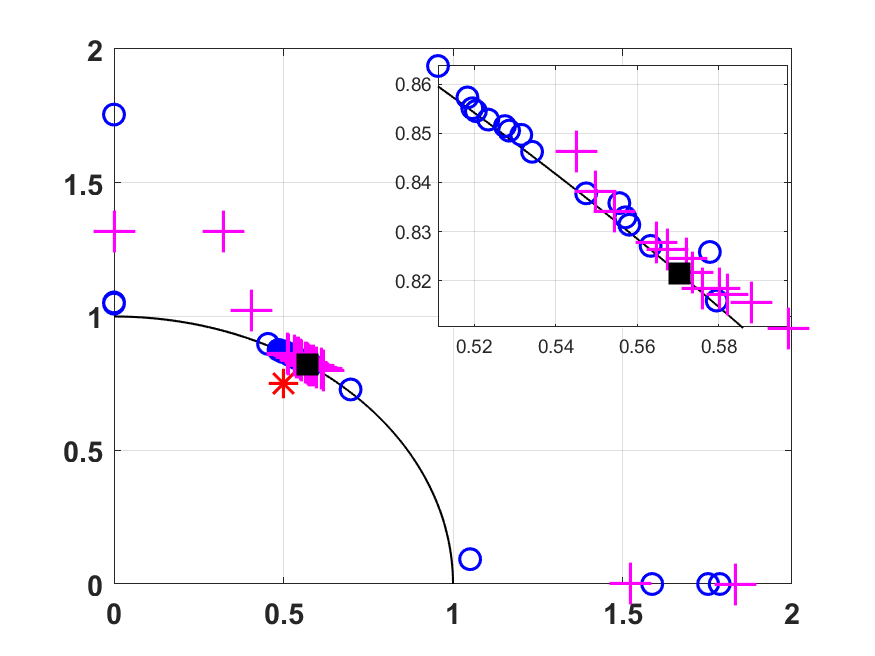}
    \caption{Performance on DTLZ5 problem}
    \label{fig:dtlz5}
\end{figure*}

\begin{figure*}[t]
    \centering
    \includegraphics[width = 0.33\textwidth]{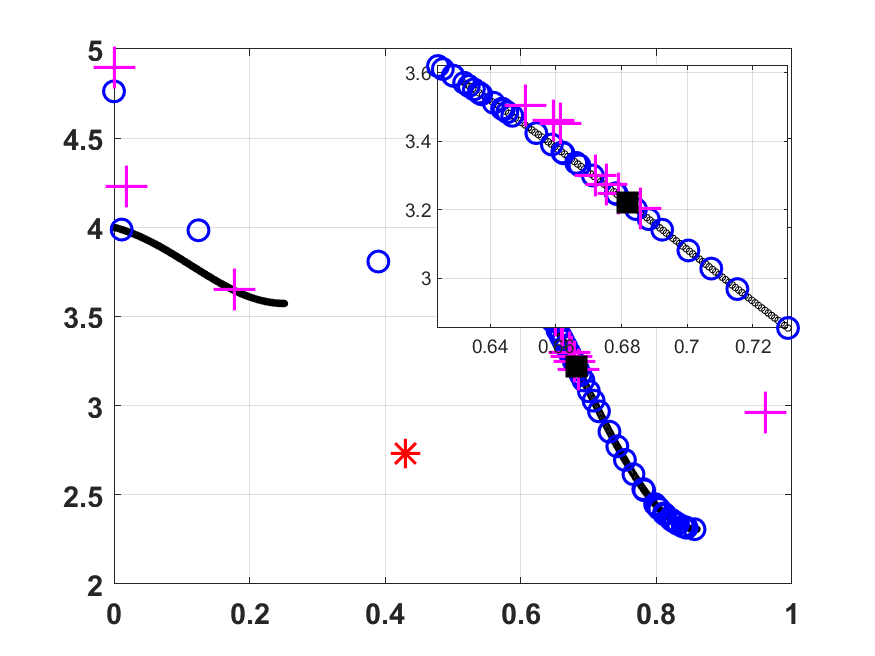}
    \includegraphics[width = 0.33\textwidth]{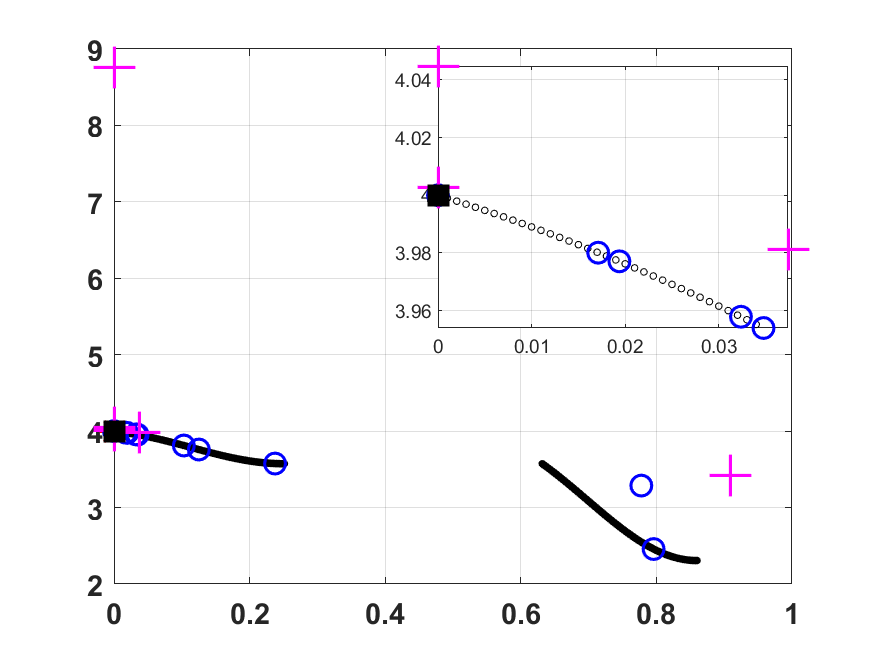}
    \includegraphics[width = 0.33\textwidth]{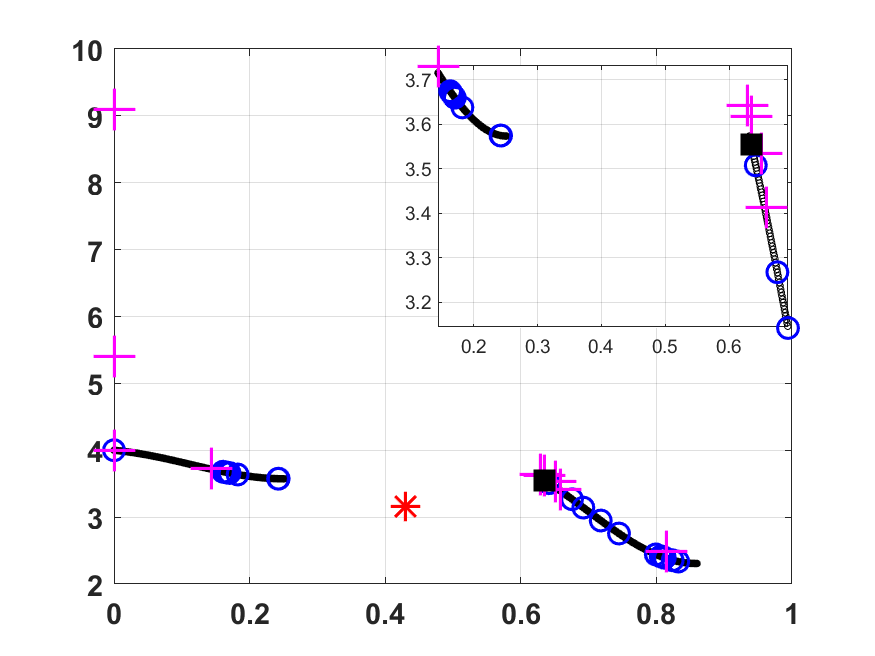}
    \includegraphics[width = 0.33\textwidth]{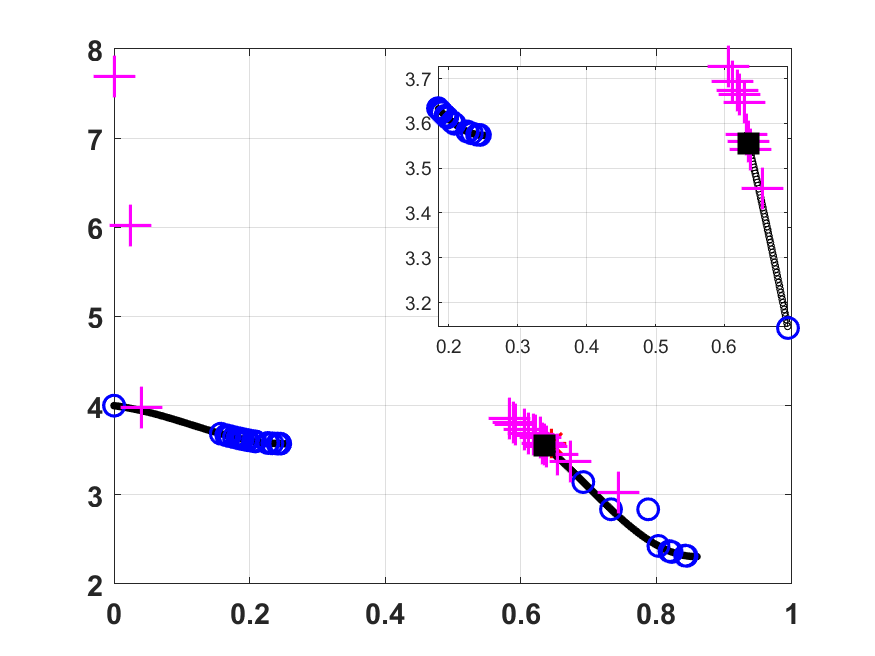}
    \includegraphics[width = 0.33\textwidth]{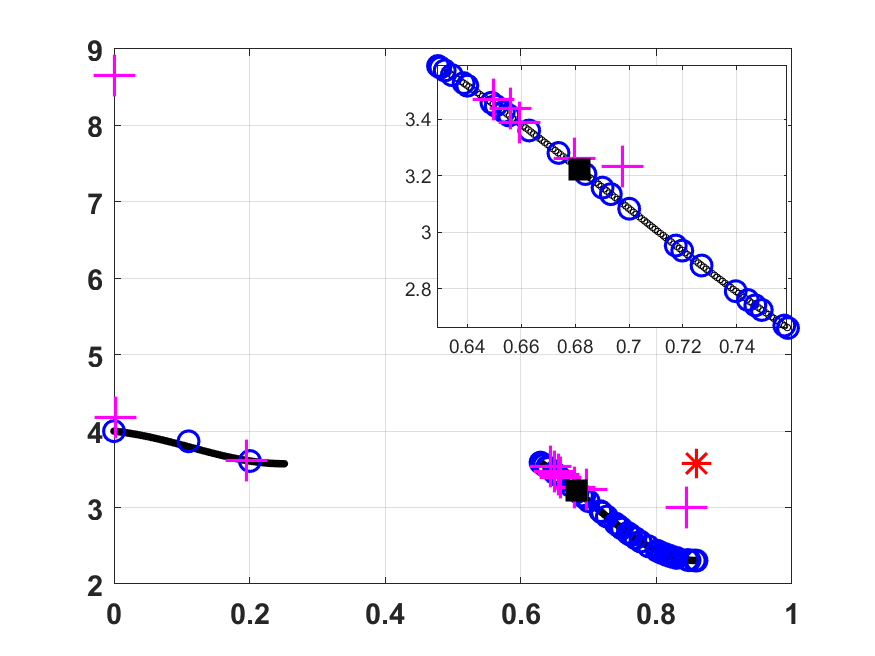}
    \includegraphics[width = 0.33\textwidth]{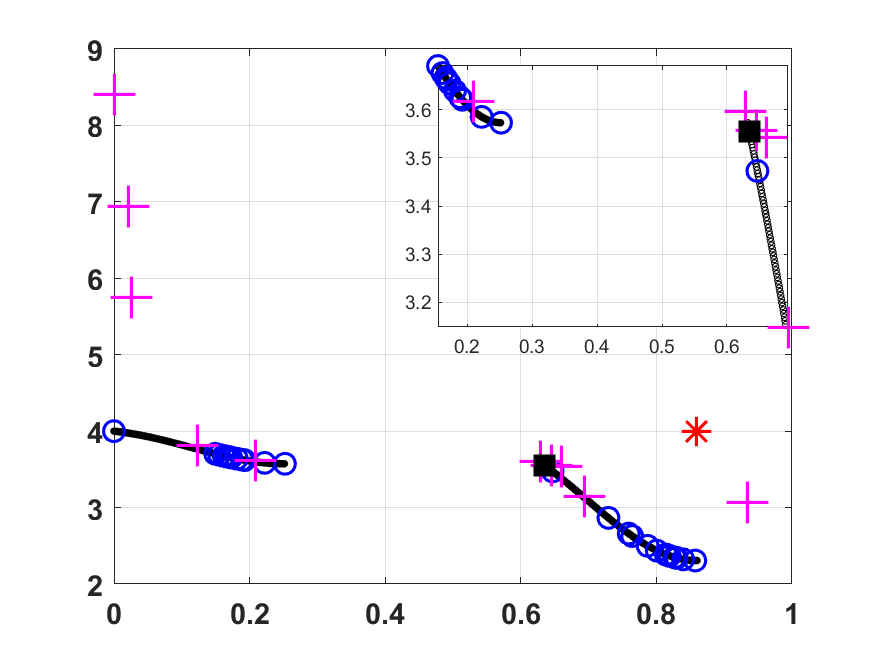}
    \caption{Performance on DTLZ7 problem}
    \label{fig:dtlz7}
\end{figure*}

For visualisation, we used scatter plots to show the solutions obtained with both the approaches. We show six instances for DTLZ problems in Figures \ref{fig:dtlz2}, \ref{fig:dtlz5} and \ref{fig:dtlz7} and the plots of the remaining instances are in the supplementary material. In the figures, we can see the reference point (or desirable objective function vector), Pareto front, nondominated solutions with mono and multi-surrogate approaches and `Ref Solution'. We did not consider the initial population when doing the nondominated sorting. From the scatter plots, it is clear that the solutions of the multi-surrogate approach converged closer to `Ref Solution' than the solutions of the mono-surrogate approach.

\begin{figure}[t]
    \centering
    \includegraphics[height = 3.9cm, width = 8.5cm]{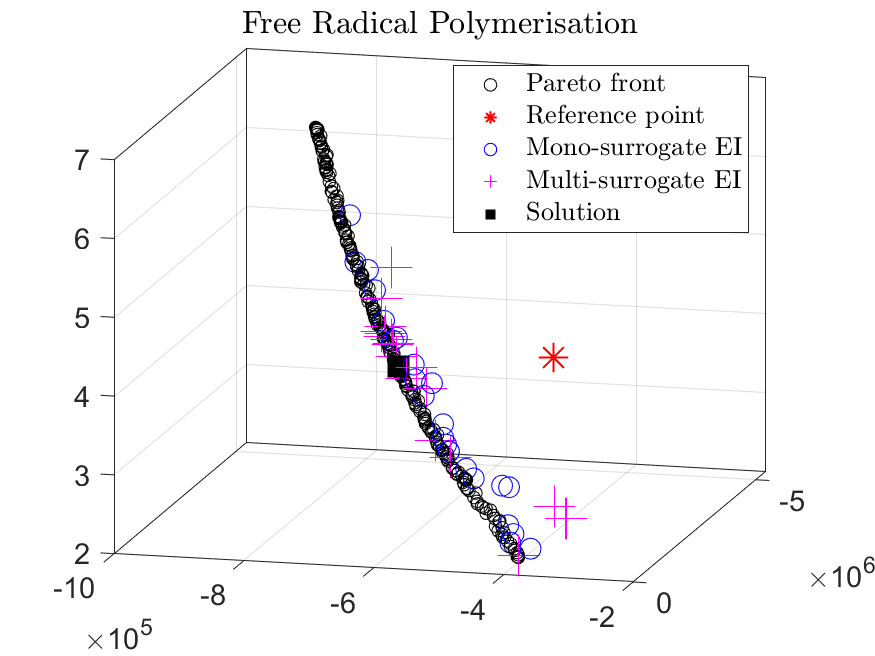}
    \includegraphics[height = 5cm, width = 8.5cm]{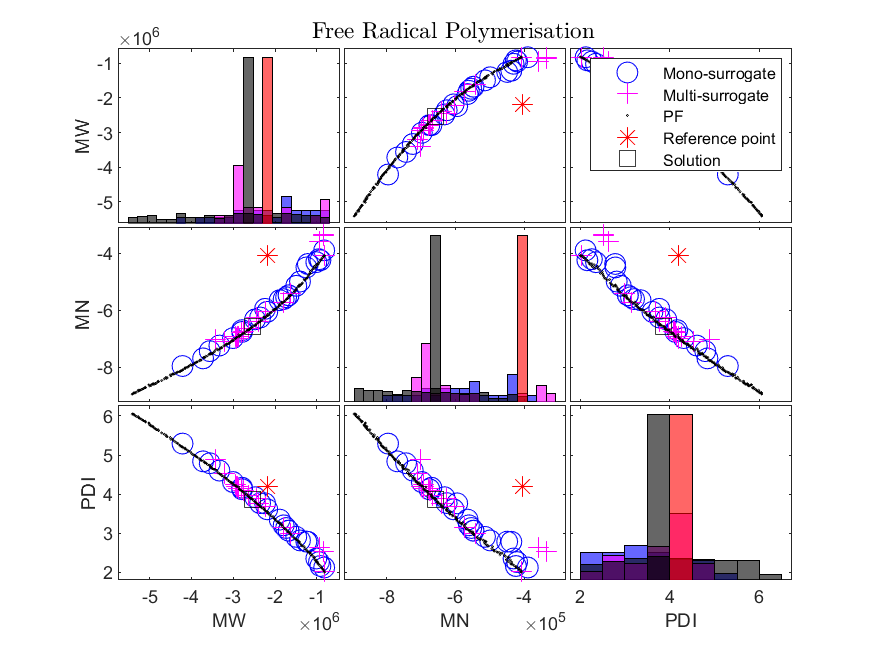}
    \caption{Performance on Free-radical Polymerisation problem}
    \label{fig:frp}
\end{figure}
\begin{figure}
    \centering
    \includegraphics[height = 4cm, width = 8.5cm]{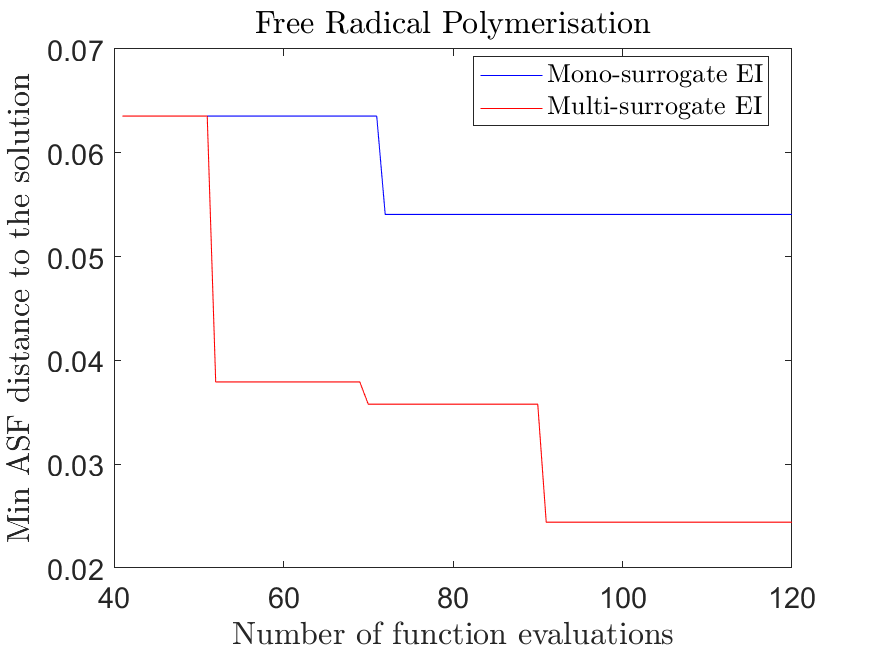}
    \caption{Minimum ASF distance to `Ref Solution' with the number of function evaluations.}
    \label{fig:frp_dist}
\end{figure}

The real-world problem is Free-radical Polymerisation problem. The problem is the manufacturing of Polyvinyl Acetate polymer. The polymer is manufactured in a batch reactor and the process can be modelled with a series of ordinary differential equations. These equations are stiff and therefore solving these equations can be computationally expensive. The computation time varies from 30 seconds to 20 minutes for one evaluation. The problem involves four decision variables: monomer concentration, initiator concentration, the temperature of the rector and polymerisation time and three objectives: maximise weight average molecular weight (MW), number average molecular weight (MN) and minimise polydispersity index (PDI). An arbitrary reference point was selected and used in both approaches. A three dimensional scatter plot between different objectives is shown in Figure \ref{fig:frp}. As can be seen, the solutions of the mono-surrogate approach are widely spread compared to the solutions of the multi-surrogate approach. The same results can also be seen in two dimensional scatter plots between different objectives in the figure. The distance between solutions of both approaches and `Ref Solution' is shown in Figure \ref{fig:frp_dist} and confirms the advantage of using multi-surrogate approach.

\begin{figure}
    \centering
    \includegraphics[height = 4cm, width = 8.5cm]{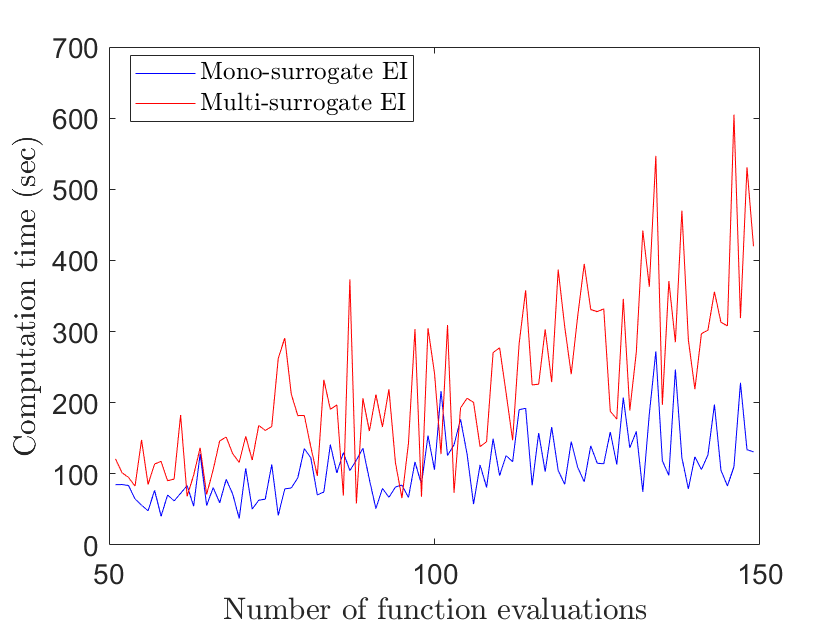}
    \caption{Computation time of mono- and multi-surrogate approaches on the DTLZ2 problem with two objectives and five decision variables}
    \label{fig:computation_time}
\end{figure}

The multi-surrogate approach relies on approximations and Monte Carlo simulations when estimating EI, which makes the optimisation process slower. The computation time on the DTLZ2 problem of one run with two objectives and five variables of both approaches is shown in Figure \ref{fig:computation_time}. As can be seen, the multi-surrogate approach is slower than the mono-surrogate approach. However, this computation time may not be significant in many real-world applications and may be negligible compared to the objective function evaluation. 

\section{Conclusions}
We presented R-MBO, a multi-surrogate approach to handle decision maker's preferences in multi-objective Bayesian optimisation. The proposed approach built independent Gaussian process models on the objective functions and approximated the distribution of the achievement scalarising function using Generalised Value Theory. We showed that the achievement scalarising function with reference point (or desirable objective function values to the decision maker) is not Gaussian distributed. After estimating the distribution, we used Monte Carlo to estimate the expected improvement as the acquisition function. We tested the approach on benchmark and real-world problems. The proofs and analysis of the results showed the potential of the approach in handling preference of the decision-maker as a-priori. 

Future work will include testing on benchmark and real-world problems with different number of objectives and decision variables. As mentioned, the multi-surrogate approach is computationally expensive. An alternative to alleviate the computation cost is to use the Laplace approximation of the approximated distribution of the scalarising function. We provided initial calculations of the Laplace approximation in the supplementary material. We plan to work on using these calculations in the future.

\color{black}

\bibliographystyle{ACM-Reference-Format}
\bibliography{main}


\begin{thebibliography}{30}


\ifx \showCODEN    \undefined \def \showCODEN     #1{\unskip}     \fi
\ifx \showDOI      \undefined \def \showDOI       #1{#1}\fi
\ifx \showISBNx    \undefined \def \showISBNx     #1{\unskip}     \fi
\ifx \showISBNxiii \undefined \def \showISBNxiii  #1{\unskip}     \fi
\ifx \showISSN     \undefined \def \showISSN      #1{\unskip}     \fi
\ifx \showLCCN     \undefined \def \showLCCN      #1{\unskip}     \fi
\ifx \shownote     \undefined \def \shownote      #1{#1}          \fi
\ifx \showarticletitle \undefined \def \showarticletitle #1{#1}   \fi
\ifx \showURL      \undefined \def \showURL       {\relax}        \fi
\providecommand\bibfield[2]{#2}
\providecommand\bibinfo[2]{#2}
\providecommand\natexlab[1]{#1}
\providecommand\showeprint[2][]{arXiv:#2}

\bibitem[\protect\citeauthoryear{Azzalini}{Azzalini}{1985}]%
        {Azzalini1985}
\bibfield{author}{\bibinfo{person}{A. Azzalini}.}
  \bibinfo{year}{1985}\natexlab{}.
\newblock \showarticletitle{A Class of Distributions Which Includes the Normal
  Ones}.
\newblock \bibinfo{journal}{\emph{Scandinavian Journal of Statistics}}
  \bibinfo{volume}{12}, \bibinfo{number}{2} (\bibinfo{year}{1985}),
  \bibinfo{pages}{171--178}.
\newblock


\bibitem[\protect\citeauthoryear{de~Haan and Ferreira}{de~Haan and
  Ferreira}{2006}]%
        {Haan2006}
\bibfield{author}{\bibinfo{person}{Laurens de Haan} {and} \bibinfo{person}{Ana
  Ferreira}.} \bibinfo{year}{2006}\natexlab{}.
\newblock \bibinfo{booktitle}{\emph{Extreme Value Theory}}.
\newblock \bibinfo{publisher}{Springer New York}.
\newblock
\urldef\tempurl%
\url{https://doi.org/10.1007/0-387-34471-3}
\showDOI{\tempurl}


\bibitem[\protect\citeauthoryear{Deb}{Deb}{2000}]%
        {Deb2000}
\bibfield{author}{\bibinfo{person}{K. Deb}.} \bibinfo{year}{2000}\natexlab{}.
\newblock \showarticletitle{An efficient constraint handling method for genetic
  algorithms}.
\newblock \bibinfo{journal}{\emph{Computer Methods in Applied Mechanics and
  Engineering}} \bibinfo{volume}{186}, \bibinfo{number}{2}
  (\bibinfo{year}{2000}), \bibinfo{pages}{311--338}.
\newblock


\bibitem[\protect\citeauthoryear{Deb, Miettinen, and Chaudhuri}{Deb
  et~al\mbox{.}}{2010}]%
        {Deb2010}
\bibfield{author}{\bibinfo{person}{K. Deb}, \bibinfo{person}{K. Miettinen},
  {and} \bibinfo{person}{S. Chaudhuri}.} \bibinfo{year}{2010}\natexlab{}.
\newblock \showarticletitle{Toward an estimation of nadir objective vector
  using a hybrid of evolutionary and local search approaches}.
\newblock \bibinfo{journal}{\emph{IEEE Transactions on Evolutionary
  Computation}}  \bibinfo{volume}{6} (\bibinfo{year}{2010}),
  \bibinfo{pages}{821--841}.
\newblock
\showISSN{0009-2509}


\bibitem[\protect\citeauthoryear{Deb, Prarap, Agarwal, and Meyarivan}{Deb
  et~al\mbox{.}}{2002}]%
        {Deb2002}
\bibfield{author}{\bibinfo{person}{K. Deb}, \bibinfo{person}{A. Prarap},
  \bibinfo{person}{S. Agarwal}, {and} \bibinfo{person}{T. Meyarivan}.}
  \bibinfo{year}{2002}\natexlab{}.
\newblock \showarticletitle{A fast and elitist multiobjective genetic
  algorithm: {NSGA-II}}.
\newblock \bibinfo{journal}{\emph{IEEE Transactions on Evolutionary
  Computation}}  \bibinfo{volume}{6} (\bibinfo{year}{2002}),
  \bibinfo{pages}{182--197}.
\newblock
\showISSN{1089-778X}


\bibitem[\protect\citeauthoryear{Deb, Thiele, Laumanns, and Zitzler}{Deb
  et~al\mbox{.}}{2005}]%
        {Deb2005b}
\bibfield{author}{\bibinfo{person}{Kalyanmoy Deb}, \bibinfo{person}{Lothar
  Thiele}, \bibinfo{person}{Marco Laumanns}, {and} \bibinfo{person}{Eckart
  Zitzler}.} \bibinfo{year}{2005}\natexlab{}.
\newblock \bibinfo{booktitle}{\emph{Scalable Test Problems for Evolutionary
  Multiobjective Optimization}}.
\newblock \bibinfo{publisher}{Springer London}, \bibinfo{address}{London},
  \bibinfo{pages}{105--145}.
\newblock


\bibitem[\protect\citeauthoryear{Emmerich, Deutz, and Klinkenberg}{Emmerich
  et~al\mbox{.}}{2011}]%
        {Emmerich2011}
\bibfield{author}{\bibinfo{person}{M. Emmerich}, \bibinfo{person}{A.H. Deutz},
  {and} \bibinfo{person}{J.W. Klinkenberg}.} \bibinfo{year}{2011}\natexlab{}.
\newblock \showarticletitle{Hypervolume-based expected improvement:
  Monotonicity properties and exact computation}. In
  \bibinfo{booktitle}{\emph{Proceedings of the IEEE Congress on Evolutionary
  Computation}}. \bibinfo{publisher}{IEEE}, \bibinfo{pages}{2147--2154}.
\newblock


\bibitem[\protect\citeauthoryear{Emmerich, Giannakoglou, and Naujoks}{Emmerich
  et~al\mbox{.}}{2006}]%
        {Emmerich2006}
\bibfield{author}{\bibinfo{person}{M. Emmerich}, \bibinfo{person}{K.
  Giannakoglou}, {and} \bibinfo{person}{B. Naujoks}.}
  \bibinfo{year}{2006}\natexlab{}.
\newblock \showarticletitle{Single- and multiobjective evolutionary
  optimization assisted by {G}aussian random field metamodels}.
\newblock \bibinfo{journal}{\emph{IEEE Transactions on Evolutionary
  Computation}}  \bibinfo{volume}{10} (\bibinfo{year}{2006}),
  \bibinfo{pages}{421--439}.
\newblock
\showISSN{1089-778X}


\bibitem[\protect\citeauthoryear{Emmerich and Naujoks}{Emmerich and
  Naujoks}{2004}]%
        {Emmerich2004}
\bibfield{author}{\bibinfo{person}{M. Emmerich} {and} \bibinfo{person}{B.
  Naujoks}.} \bibinfo{year}{2004}\natexlab{}.
\newblock \showarticletitle{Metamodel assisted multiobjective optimisation
  strategies and their application in airfoil design}.
\newblock In \bibinfo{booktitle}{\emph{Adaptive Computing in Design and
  Manufacture VI}}, \bibfield{editor}{\bibinfo{person}{I.~C. Parmee}} (Ed.).
  \bibinfo{publisher}{Springer London}, \bibinfo{pages}{249--260}.
\newblock
\showISBNx{978-0-85729-338-1}


\bibitem[\protect\citeauthoryear{Emmerich}{Emmerich}{2005}]%
        {Emmerich2005a}
\bibfield{author}{\bibinfo{person}{Michael T.~M. Emmerich}.}
  \bibinfo{year}{2005}\natexlab{}.
\newblock \showarticletitle{Single- and multi-objective evolutionary design
  optimization assisted by gaussian random field metamodels}.
\newblock  (\bibinfo{year}{2005}).
\newblock
\urldef\tempurl%
\url{https://doi.org/10.17877/DE290R-57}
\showDOI{\tempurl}


\bibitem[\protect\citeauthoryear{Hakanen and Knowles}{Hakanen and
  Knowles}{2017}]%
        {Hakanen2017}
\bibfield{author}{\bibinfo{person}{Jussi Hakanen} {and}
  \bibinfo{person}{Joshua~D. Knowles}.} \bibinfo{year}{2017}\natexlab{}.
\newblock \showarticletitle{On Using Decision Maker Preferences with ParEGO}.
  In \bibinfo{booktitle}{\emph{9th International Conference on Evolutionary
  Multi-Criterion Optimization - Volume 10173}} (M\"{u}nster, Germany)
  \emph{(\bibinfo{series}{EMO 2017})}. \bibinfo{publisher}{Springer-Verlag},
  \bibinfo{address}{Berlin, Heidelberg}, \bibinfo{pages}{282--297}.
\newblock
\showISBNx{9783319541563}
\urldef\tempurl%
\url{https://doi.org/10.1007/978-3-319-54157-0_20}
\showDOI{\tempurl}


\bibitem[\protect\citeauthoryear{Knowles}{Knowles}{2006}]%
        {Knowles2006}
\bibfield{author}{\bibinfo{person}{J. Knowles}.}
  \bibinfo{year}{2006}\natexlab{}.
\newblock \showarticletitle{Par{EGO}: a hybrid algorithm with on-line landscape
  approximation for expensive multiobjective optimization problems}.
\newblock \bibinfo{journal}{\emph{IEEE Transactions on Evolutionary
  Computation}}  \bibinfo{volume}{10} (\bibinfo{year}{2006}),
  \bibinfo{pages}{50--66}.
\newblock


\bibitem[\protect\citeauthoryear{Lemons}{Lemons}{2002}]%
        {Lemons2002}
\bibfield{author}{\bibinfo{person}{Don Lemons}.}
  \bibinfo{year}{2002}\natexlab{}.
\newblock \bibinfo{booktitle}{\emph{An introduction to stochastic processes in
  physics : containing "On the theory of Brownian motion" by Paul Langevin,
  translated by Anthony Gythiel}}.
\newblock \bibinfo{publisher}{Johns Hopkins University Press},
  \bibinfo{address}{Baltimore}.
\newblock


\bibitem[\protect\citeauthoryear{Mazumdar, Chugh, Hakanen, and
  Miettinen}{Mazumdar et~al\mbox{.}}{2022}]%
        {Atanu_TEVC}
\bibfield{author}{\bibinfo{person}{Atanu Mazumdar}, \bibinfo{person}{Tinkle
  Chugh}, \bibinfo{person}{Jussi Hakanen}, {and} \bibinfo{person}{Kaisa
  Miettinen}.} \bibinfo{year}{2022}\natexlab{}.
\newblock \showarticletitle{Probabilistic Selection Approaches in
  Decomposition-based Evolutionary Algorithms for Offline Data-Driven
  Multiobjective Optimization}.
\newblock \bibinfo{journal}{\emph{IEEE Transactions on Evolutionary
  Computation}} (\bibinfo{year}{2022}), \bibinfo{pages}{1--1}.
\newblock
\urldef\tempurl%
\url{https://doi.org/10.1109/TEVC.2022.3154231}
\showDOI{\tempurl}


\bibitem[\protect\citeauthoryear{Mckay, Beckman, and Conover}{Mckay
  et~al\mbox{.}}{2000}]%
        {Mckay2000}
\bibfield{author}{\bibinfo{person}{M.D. Mckay}, \bibinfo{person}{R.J. Beckman},
  {and} \bibinfo{person}{W.J. Conover}.} \bibinfo{year}{2000}\natexlab{}.
\newblock \showarticletitle{A comparison of three methods for selecting values
  of input variables in the analysis of output from a computer code}.
\newblock \bibinfo{journal}{\emph{Technometrics}}  \bibinfo{volume}{42}
  (\bibinfo{year}{2000}), \bibinfo{pages}{55--61}.
\newblock


\bibitem[\protect\citeauthoryear{Miettinen}{Miettinen}{1999}]%
        {Miettinen1999}
\bibfield{author}{\bibinfo{person}{K. Miettinen}.}
  \bibinfo{year}{1999}\natexlab{}.
\newblock \bibinfo{booktitle}{\emph{Nonlinear multiobjective optimization}}.
\newblock \bibinfo{publisher}{Kluwer}, \bibinfo{address}{Boston, MA}.
\newblock


\bibitem[\protect\citeauthoryear{Miettinen, Hakanen, and Podkopaev}{Miettinen
  et~al\mbox{.}}{2016}]%
        {Miettinen2016}
\bibfield{author}{\bibinfo{person}{K. Miettinen}, \bibinfo{person}{J. Hakanen},
  {and} \bibinfo{person}{D. Podkopaev}.} \bibinfo{year}{2016}\natexlab{}.
\newblock \showarticletitle{Interactive nonlinear multiobjective optimization
  methods}.
\newblock In \bibinfo{booktitle}{\emph{Multiple Criteria Decision Analysis:
  State of the Art Surveys}}, \bibfield{editor}{\bibinfo{person}{S.~Greco},
  \bibinfo{person}{M.~Ehrgott}, {and} \bibinfo{person}{J.R. Figueira}} (Eds.).
  \bibinfo{publisher}{Springer New York}, \bibinfo{pages}{927--976}.
\newblock
\showISBNx{978-1-4939-3094-4}


\bibitem[\protect\citeauthoryear{Miettinen and M{\"a}kel{\"a}}{Miettinen and
  M{\"a}kel{\"a}}{2002}]%
        {Miettinen2002}
\bibfield{author}{\bibinfo{person}{Kaisa Miettinen} {and}
  \bibinfo{person}{Marko~M. M{\"a}kel{\"a}}.} \bibinfo{year}{2002}\natexlab{}.
\newblock \showarticletitle{On scalarizing functions in multiobjective
  optimization}.
\newblock \bibinfo{journal}{\emph{OR Spectrum}} \bibinfo{volume}{24},
  \bibinfo{number}{2} (\bibinfo{date}{01 May} \bibinfo{year}{2002}),
  \bibinfo{pages}{193--213}.
\newblock
\showISSN{1436-6304}
\urldef\tempurl%
\url{https://doi.org/10.1007/s00291-001-0092-9}
\showDOI{\tempurl}


\bibitem[\protect\citeauthoryear{Miettinen, Ruiz, and Wierzbicki}{Miettinen
  et~al\mbox{.}}{2008}]%
        {Miettinen2008}
\bibfield{author}{\bibinfo{person}{K. Miettinen}, \bibinfo{person}{F. Ruiz},
  {and} \bibinfo{person}{A.P. Wierzbicki}.} \bibinfo{year}{2008}\natexlab{}.
\newblock \showarticletitle{Introduction to Multiobjective Optimization:
  {I}nteractive Approaches}.
\newblock In \bibinfo{booktitle}{\emph{Multiobjective Optimization:
  {I}nteractive and Evolutionary Approaches}},
  \bibfield{editor}{\bibinfo{person}{J.~Branke}, \bibinfo{person}{K.~Deb},
  \bibinfo{person}{K.~Miettinen}, {and}
  \bibinfo{person}{R.~S{\l}owi{\'{n}}ski}} (Eds.). \bibinfo{publisher}{Springer
  Berlin Heidelberg}, \bibinfo{pages}{27--57}.
\newblock
\showISBNx{978-3-540-88908-3}


\bibitem[\protect\citeauthoryear{Mogilicharla, Chugh, Majumder, and
  Mitra}{Mogilicharla et~al\mbox{.}}{2014}]%
        {Chugh2014}
\bibfield{author}{\bibinfo{person}{A. Mogilicharla}, \bibinfo{person}{T.
  Chugh}, \bibinfo{person}{S. Majumder}, {and} \bibinfo{person}{K. Mitra}.}
  \bibinfo{year}{2014}\natexlab{}.
\newblock \showarticletitle{Multi-objective optimization of bulk Vinyl Acetate
  polymerization with branching}.
\newblock \bibinfo{journal}{\emph{Materials and Manufacturing Processes}}
  \bibinfo{volume}{29} (\bibinfo{year}{2014}), \bibinfo{pages}{210--217}.
\newblock


\bibitem[\protect\citeauthoryear{Mood}{Mood}{1974}]%
        {Mood1974}
\bibfield{author}{\bibinfo{person}{Alexander Mood}.}
  \bibinfo{year}{1974}\natexlab{}.
\newblock \bibinfo{booktitle}{\emph{Introduction to the theory of statistics}}.
\newblock \bibinfo{publisher}{McGraw-Hill}, \bibinfo{address}{Auckland}.
\newblock


\bibitem[\protect\citeauthoryear{Nadarajah and Kotz}{Nadarajah and
  Kotz}{2008}]%
        {Nadarajah2008}
\bibfield{author}{\bibinfo{person}{Saralees Nadarajah} {and}
  \bibinfo{person}{Samuel Kotz}.} \bibinfo{year}{2008}\natexlab{}.
\newblock \showarticletitle{Exact Distribution of the Max/Min of Two Gaussian
  Random Variables}.
\newblock \bibinfo{journal}{\emph{IEEE Transactions on Very Large Scale
  Integration (VLSI) Systems}} \bibinfo{volume}{16}, \bibinfo{number}{2}
  (\bibinfo{year}{2008}), \bibinfo{pages}{210--212}.
\newblock


\bibitem[\protect\citeauthoryear{Palar, Zuhal, Chugh, and Rahat}{Palar
  et~al\mbox{.}}{2020}]%
        {Palar_AIAA}
\bibfield{author}{\bibinfo{person}{P.~S. Palar}, \bibinfo{person}{L.~R. Zuhal},
  \bibinfo{person}{T. Chugh}, {and} \bibinfo{person}{A. Rahat}.}
  \bibinfo{year}{2020}\natexlab{}.
\newblock \showarticletitle{On the Impact of Covariance Functions in
  Multi-Objective Bayesian Optimization for Engineering Design}. In
  \bibinfo{booktitle}{\emph{AIAA Scitech 2020 Forum}}.
\newblock
\urldef\tempurl%
\url{https://doi.org/10.2514/6.2020-1867}
\showDOI{\tempurl}
\showeprint{https://arc.aiaa.org/doi/pdf/10.2514/6.2020-1867}


\bibitem[\protect\citeauthoryear{Rahat, Everson, and Fieldsend}{Rahat
  et~al\mbox{.}}{2017}]%
        {Alma2017}
\bibfield{author}{\bibinfo{person}{Alma A.~M. Rahat},
  \bibinfo{person}{Richard~M. Everson}, {and} \bibinfo{person}{Jonathan~E.
  Fieldsend}.} \bibinfo{year}{2017}\natexlab{}.
\newblock \showarticletitle{Alternative Infill Strategies for Expensive
  Multi-objective Optimisation}. In \bibinfo{booktitle}{\emph{Proceedings of
  the Genetic and Evolutionary Computation Conference}} (Berlin, Germany)
  \emph{(\bibinfo{series}{GECCO '17})}. \bibinfo{publisher}{ACM},
  \bibinfo{address}{New York, NY, USA}, \bibinfo{pages}{873--880}.
\newblock
\showISBNx{978-1-4503-4920-8}
\urldef\tempurl%
\url{https://doi.org/10.1145/3071178.3071276}
\showDOI{\tempurl}


\bibitem[\protect\citeauthoryear{Rasmussen and Williams}{Rasmussen and
  Williams}{2006}]%
        {Rasmussen2006}
\bibfield{author}{\bibinfo{person}{C.~E. Rasmussen} {and}
  \bibinfo{person}{C.~K.~I. Williams}.} \bibinfo{year}{2006}\natexlab{}.
\newblock \bibinfo{booktitle}{\emph{Gaussian processes for machine learning}}.
\newblock \bibinfo{publisher}{The MIT Press}.
\newblock
\showISBNx{0-262-18253-X}


\bibitem[\protect\citeauthoryear{T.Chugh, Rahat, and Palar}{T.Chugh
  et~al\mbox{.}}{2019}]%
        {Chugh_2019_LOD}
\bibfield{author}{\bibinfo{person}{T.Chugh}, \bibinfo{person}{A. Rahat}, {and}
  \bibinfo{person}{P.S. Palar}.} \bibinfo{year}{2019}\natexlab{}.
\newblock \showarticletitle{Trading-off Data Fit and Complexity in Training
  Gaussian Processes with Multiple Kernels}. In
  \bibinfo{booktitle}{\emph{Proceedings of the International Conference on
  Machine Learning, Optimization, and Data Science (LOD)}}
  \emph{(\bibinfo{series}{Lecture Notes in Computer Science (LNCS)})}.
  \bibinfo{publisher}{Springer}.
\newblock


\bibitem[\protect\citeauthoryear{Wang and Jegelka}{Wang and Jegelka}{2017}]%
        {GumbelBO}
\bibfield{author}{\bibinfo{person}{Zi Wang} {and} \bibinfo{person}{Stefanie
  Jegelka}.} \bibinfo{year}{2017}\natexlab{}.
\newblock \showarticletitle{Max-Value Entropy Search for Efficient Bayesian
  Optimization}. In \bibinfo{booktitle}{\emph{Proceedings of the 34th
  International Conference on Machine Learning - Volume 70}} (Sydney, NSW,
  Australia) \emph{(\bibinfo{series}{ICML'17})}. \bibinfo{publisher}{JMLR.org},
  \bibinfo{pages}{3627–3635}.
\newblock


\bibitem[\protect\citeauthoryear{Weisstein}{Weisstein}{[n.\,d.]}]%
        {GEV}
\bibfield{author}{\bibinfo{person}{Eric~W Weisstein}.}
  \bibinfo{year}{[n.\,d.]}\natexlab{}.
\newblock \bibinfo{howpublished}{Weisstein, Eric W. "Extreme Value
  Distribution". mathworld.wolfram.com.}.
\newblock


\bibitem[\protect\citeauthoryear{Wierzbicki}{Wierzbicki}{1980}]%
        {Wierzbicki1980}
\bibfield{author}{\bibinfo{person}{A.P. Wierzbicki}.}
  \bibinfo{year}{1980}\natexlab{}.
\newblock \showarticletitle{The use of reference objectives in multiobjective
  optimization}. In \bibinfo{booktitle}{\emph{Proceedings of the Multiple
  Criteria Decision Making Theory and Application}},
  \bibfield{editor}{\bibinfo{person}{G.~Fandel} {and} \bibinfo{person}{T.~Gal}}
  (Eds.). \bibinfo{publisher}{Springer}, \bibinfo{pages}{468--486}.
\newblock


\bibitem[\protect\citeauthoryear{Yang, Emmerich, Deutz, and Bäck}{Yang
  et~al\mbox{.}}{2019}]%
        {Yang2019}
\bibfield{author}{\bibinfo{person}{Kaifeng Yang}, \bibinfo{person}{Michael
  Emmerich}, \bibinfo{person}{Andr{\'{e}} Deutz}, {and} \bibinfo{person}{Thomas
  Bäck}.} \bibinfo{year}{2019}\natexlab{}.
\newblock \showarticletitle{Multi-Objective Bayesian Global Optimization using
  expected hypervolume improvement gradient}.
\newblock \bibinfo{journal}{\emph{Swarm and Evolutionary Computation}}
  \bibinfo{volume}{44} (\bibinfo{date}{feb} \bibinfo{year}{2019}),
  \bibinfo{pages}{945--956}.
\newblock
\urldef\tempurl%
\url{https://doi.org/10.1016/j.swevo.2018.10.007}
\showDOI{\tempurl}


\end{thebibliography}
\clearpage
\includepdf[pages={1-10}]{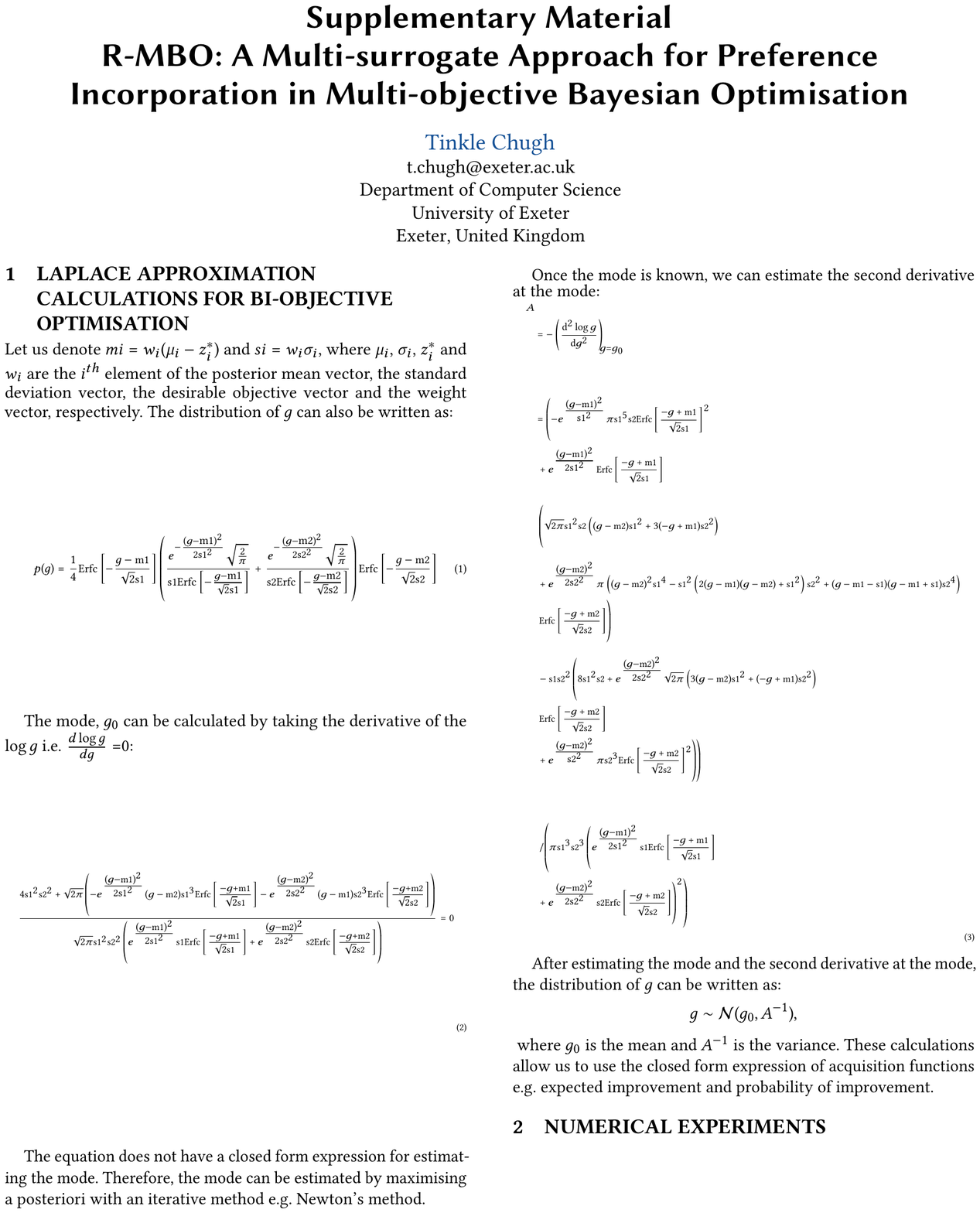}

\end{document}